\documentclass{article}
\usepackage{microtype}
\usepackage{graphicx}
\usepackage{subcaption}
\usepackage{booktabs}
\usepackage{hyperref}

\usepackage[preprint]{icml2026}
\usepackage{amsmath}
\usepackage{amssymb}
\usepackage{mathtools}
\usepackage{amsthm}
\usepackage[capitalize,noabbrev]{cleveref}
\usepackage{placeins} % for FloatBarrier
\usepackage{multirow} % multirows in tables
\usepackage{colortbl}

\usepackage[table]{xcolor}
\usepackage{siunitx}

\definecolor{HeaderBG}{HTML}{F5F6F8}
\definecolor{OursBG}{HTML}{FFF9E6}
\definecolor{DiffLow}{HTML}{2E7D32}
\definecolor{DiffMed}{HTML}{1B5E20}
\definecolor{DiffHigh}{HTML}{0B3D0B}
\definecolor{DiffNeg}{HTML}{6E6E6E}

\newcommand{\diffpos}[2]{#1\text{\scriptsize\hspace{1pt}\textcolor{DiffLow}{(+#2)}}}

\newcommand{\diffneg}[2]{#1\text{\scriptsize\hspace{1pt}\textcolor{DiffNeg}{(-#2)}}}
\newcommand{\deltapos}[1]{\text{\scriptsize\hspace{1pt}\textcolor{DiffLow}{(+#1)}}}
\newcommand{\deltaposmed}[1]{\text{\scriptsize\hspace{1pt}\textcolor{DiffMed}{(+#1)}}}
\newcommand{\deltaposhigh}[1]{\text{\scriptsize\hspace{1pt}\textcolor{DiffHigh}{(+#1)}}}
\newcommand{\deltaneg}[1]{\text{\scriptsize\hspace{1pt}\textcolor{DiffNeg}{(-#1)}}}

\usepackage{fvextra} % extends fancyvrb's Verbatim
\DefineVerbatimEnvironment{prompt}{Verbatim}{
  breaklines=true,
  breakanywhere=true,
  fontsize=\footnotesize,
  xleftmargin=2em,
  frame=single,
}
\usepackage{newfloat}
\DeclareFloatingEnvironment[
    fileext=lop,
    placement={!ht},
    name=Prompt
]{promptenv}

\usepackage{caption}
\DeclareCaptionStyle{promptstyle}{
  aboveskip=-4pt,
  belowskip=-4pt
}
\crefname{promptenv}{Prompt}{Prompts}
\Crefname{promptenv}{Prompt}{Prompts}

\icmltitlerunning{ArtifactLens: Hundreds of Labels Are Enough for Artifact Detection with VLMs}
\newcommand{\SystemName}{ArtifactLens}

\newsavebox{\pullfigbox}
\savebox{\pullfigbox}{%
  \includegraphics[page=1, width=\textwidth,clip,trim=0 610 605 0]{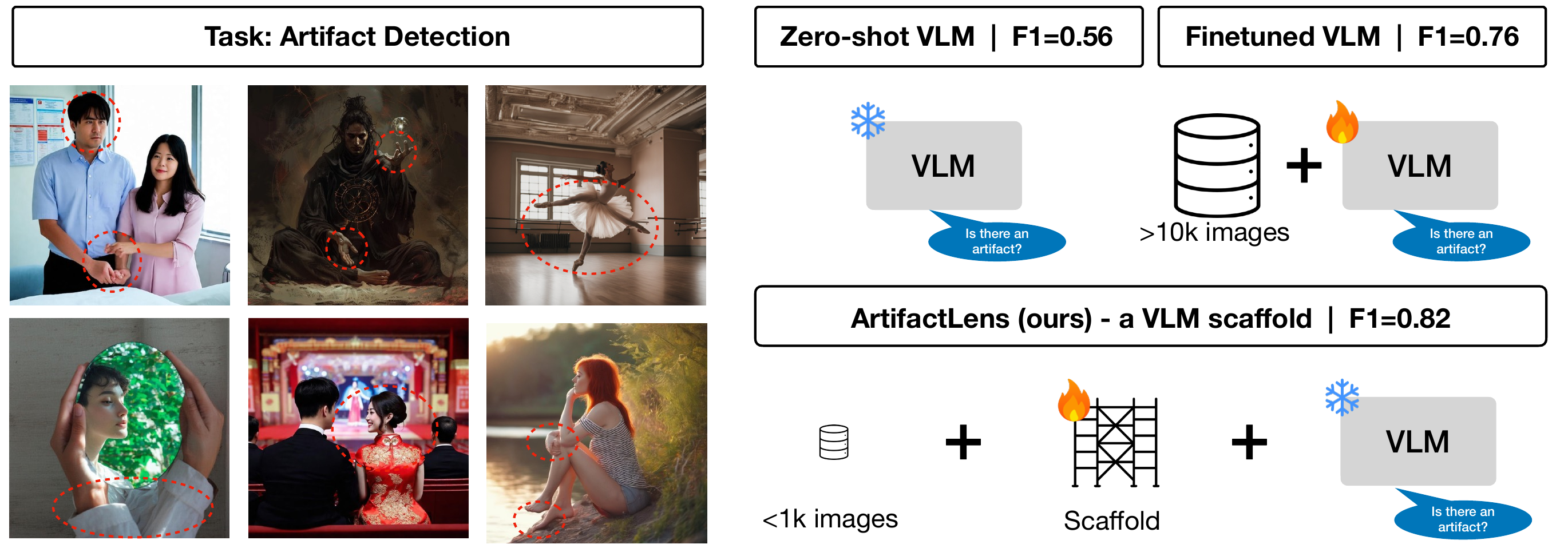}%
}
\begin{document}
\twocolumn[
  \icmltitle{ArtifactLens: Hundreds of Labels Are Enough for Artifact Detection with VLMs}
  % Author info
  \icmlsetsymbol{equal}{*}
  \begin{icmlauthorlist}
    \icmlauthor{James Burgess}{stanford}
    \icmlauthor{Rameen Abdal}{snap}
    \icmlauthor{Dan Stoddart}{snap}
    \icmlauthor{Sergey Tulyakov}{snap}
    \icmlauthor{Serena Yeung-Levy}{stanford}
    \icmlauthor{Kuan-Chieh Jackson Wang}{snap}
  \end{icmlauthorlist}
  \icmlaffiliation{stanford}{Stanford University, Stanford, CA, USA}
  \icmlaffiliation{snap}{Snap Inc., Santa Monica, CA, USA}
  \icmlcorrespondingauthor{James Burgess}{jmhb@stanford.edu}
  \icmlkeywords{Vision-Language Models, Artifact Detection, Prompt Optimization}
  \vskip 0.1in
   \centering
  { \href{http://jmhb0.github.io/ArtifactLens}{\texttt{http://jmhb0.github.io/ArtifactLens}}
   }
   \vskip 0.2in
  % Pull figure on page 1
  \centering
  \usebox{\pullfigbox}
  \vspace{-2.0em}
  \captionsetup{type=figure}
  \caption{
    \textit{Left}: Our task is to detect human anatomy artifacts in AI-generated images by labeling images as `artifact' or `not artifact'.
    \textit{Top right}: Zero-shot VLMs perform poorly (low F1 score).
    Finetuned VLMs perform better but they need large datasets.
    \textit{Bottom right}: our system \SystemName\ uses scaffolding around frozen VLMs to achieve the best performance and with much less data.
    Specifically, we have an architecture of `specialists' that call pretrained VLMs, and the VLMs are subject to black-box optimization.
  }
  \label{fig:pull}
  \vskip 0.3in
]
\printAffiliationsAndNotice{}

\begin{abstract}
Modern image generators produce strikingly realistic images, where only artifacts like distorted hands or warped objects reveal their synthetic origin. Detecting these artifacts is essential: without detection, we cannot benchmark generators or train reward models to improve them. Current detectors fine-tune VLMs on tens of thousands of labeled images, but this is expensive to repeat whenever generators evolve or new artifact types emerge. We show that pretrained VLMs already encode the knowledge needed to detect artifacts --- with the right scaffolding, this capability can be unlocked using only a few hundred labeled examples per artifact category. Our system, ArtifactLens, achieves state-of-the-art on five human artifact benchmarks (the first evaluation across multiple datasets) while requiring orders of magnitude less labeled data. The scaffolding consists of a multi-component architecture with in-context learning and text instruction optimization, with novel improvements to each. Our methods generalize to other artifact types -- object morphology, animal anatomy, and entity interactions -- and to the distinct task of AIGC detection.
\end{abstract}

    \begin{figure*}[th]
    \centering
    \includegraphics[page=2, width=\textwidth,clip,trim=0 750 605 0]{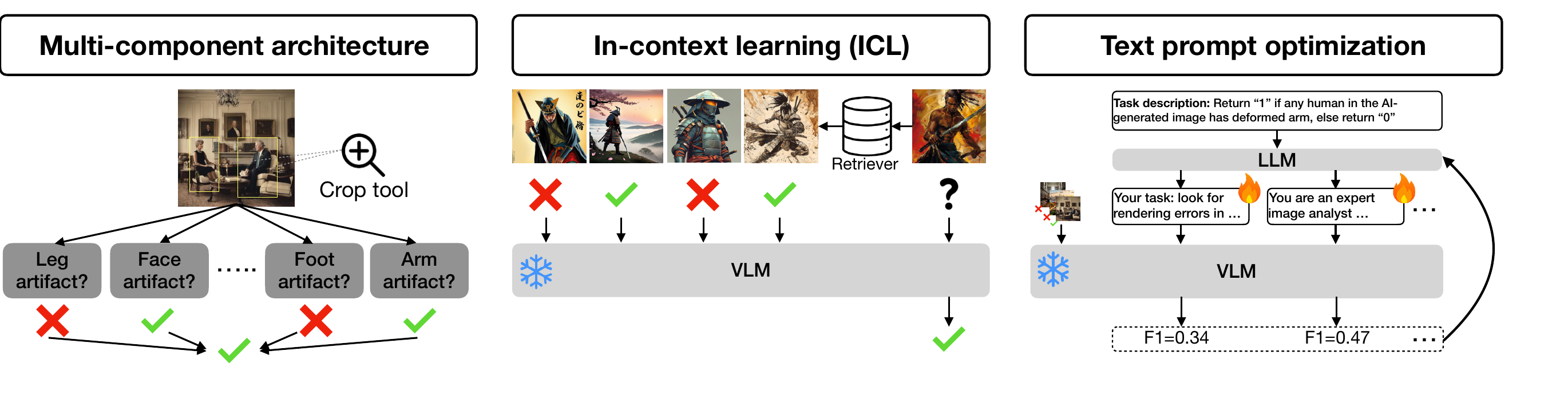}
    \captionof{figure}{
    The scaffolding methods in \SystemName.
    \textit{Left:} a multi-component architecture, where each \textit{specialist} leverages pretrained VLMs to classify a single error like `leg artifact'. The specialists use a crop tool to zoom to regions-of-interest for easier visual understanding \cite{wu2024v}. 
    \textit{Middle}: To optimize the pretrained VLMs, in-context learning uses prompts with task demonstrations, which are image-label pairs. The challenge is choosing the best task demonstrations -- our final system does retrieval-based selection.
    \textit{Right}: We also optimize the pretrained VLM with text prompt optimization. A concise seed instruction is passed to an LLM which generates candidate text prompts. The prompts are evaluated against a development dataset and the results are fed back to an LLM for 
    rewriting. 
    }
    \label{fig:methods}
\end{figure*}

% Still could include  discussion of perception vs reasoning tasks 
\section{Introduction}
\label{sec:intro}
Creative image generation has exploded in research and product as deep learning models achieve unprecedented quality and user control \cite{ramesh2022hierarchical, rombach2022high, batifol2025flux, wu2025qwen}.  Modern generators are so realistic that \textit{artifacts} are often the only giveaway that they are AI generated.
% But AI-generated content has \textit{artifacts} -- visual features that are undesirable and do not adhere to the user's intent \cite{cao2024synartifact,  fang2024humanrefiner, wang2024detecting, kang2025geneva, wang2025magicmirror, wang2025generated}. 
Human anatomy artifacts like extra limbs, distorted hands, or creepy facial expressions \cite{wang2025magicmirror, wang2024detecting, fang2024humanrefiner, wang2025generated, kang2025geneva} are the most notorious -- sometimes amusing, but sometimes disturbing, and they can limit the commercial use of AI generators. Other artifact types also present problems, such as object morphology, irrational object interactions, illegible letters, or unnatural textures \cite{cao2024synartifact, wang2025magicmirror}. 

\textit{Artifact detectors} offer a solution \cite{wang2024detecting, wang2025magicmirror}. Detectors that give image-level labels can enable benchmarking for model development; or they can be reward models used in preference finetuning \cite{fan2023dpok, wallace2024diffusion}, inference guidance \cite{kynkaanniemi2024applying, parmar2025scaling}, or dataset filtering \cite{nguyen2024swiftbrush, chen2025snapgen, wu2025lightgen}. The strongest artifact detectors finetune vision language models (VLMs) \cite{wang2025magicmirror} or object detectors \cite{fang2024humanrefiner, wang2024detecting, nguyen2024swiftbrush} (\cref{fig:pull}). However, this requires annotating large datasets (30k-300k), and the process may need repeating whenever generators evolve or new artifact types emerge. Why is artifact detection so challenging? We speculate: a lack of artifact data in VLM post-training \cite{zhang2024visually, udandarao2024no}; biases against rare classes \cite{goyal2017making, leng2024mitigating, vo2025vision}; artifacts being visually small \cite{wu2024v}; and prior works relying on smaller foundation models.

In this work, we show that pretrained VLMs can already detect artifacts -- they just need better scaffolding. Instead of finetuning VLM weights, we propose \SystemName: a multi-VLM architecture with black box optimization of its VLM components. The first key idea is that artifact detection is decomposable into multiple sub-tasks and optimized independently. We create \textit{specialists} for different error types like `deformed hands', `deformed face', or `missing leg'. The specialists can zoom to regions of interest and classify errors using a VLM that is optimized with labels for that error type -- this is a compound system architecture \cite{compound-ai-blog, khattab2024dspy} (or a workflow \cite{anthropic2024effectiveagents}). But prior works \cite{wang2025diffdoctor, wang2025magicmirror} found that zero-shot VLMs are bad artifact detectors, so how can we use them? The second key idea is that pretrained VLMs can improve enormously with black box optimization.  Specifically, we apply in-context learning (ICL) which prompts VLMs with example image-label pairs to learn the task efficiently; we retrieve demonstrations conditioned on the input image \cite{alayrac2022flamingo, doveh2024towards}. Then we apply text instruction optimization, which uses LLMs to rewrite the VLM text instruction; it searches over candidate instructions against some performance metric on a development dataset \cite{yang2023large, zhou2022large}. We implement this system in the popular DSPy framework \cite{khattab2024dspy, khattab2022demonstrate}. 

We also propose two new methods to advance the black-box optimization of LLMs and VLMS, which are growing research topics \cite{yang2023large, shinn2023reflexion, fernando2023promptbreeder, agrawal2025gepa}. In-context learning (ICL) methods prompt the VLM with demonstrations that best enable efficiently learning a task definition \cite{zong2024vl}. Our technique, \textit{counterfactual (CF) demonstrations}, groups demonstrations into pairs of similar inputs but different labels. Intuitively, VLMs will more easily learn the `deformed hand' concept given a pair of images that are semantically similar in most ways except for the `deformed hand' label. We instantiate CF demos for grouping using CLIP as a similarity metric \cite{radford2021learning}. Next, in text prompt optimization, we find that LLM instruction generators usually generate prompts encouraging the VLM to be cautious -- ``only mark artifacts when you're confident''. But VLMs are already biased against the artifact class, and they benefit from the opposite guidance.  We propose \textit{full spectrum prompting}: we instruct the LLM generator to create a diverse prompt pool with varying confidence thresholds, from conservative (``mark artifacts only with high confidence'') to aggressive (``mark artifacts even with lower confidence''), ensuring the candidate prompt pool covers a full spectrum of decision boundaries.
% These techniques improve performance on artifact detection, but we expect that they apply to other tasks as well.

We combine five benchmarks \cite{cao2024synartifact, fang2024humanrefiner, wang2024detecting, wang2025generated, wang2025magicmirror}, and show that \SystemName\ with Gemini-2.5-Pro \cite{gemini25pro} outperforms the best finetuned model by 8\% on F1 score, while using only 10\% of the data; and in fact, performance only degrades by 9\% using only 200 training samples. For the same base zero-shot VLM, the scaffolding improves all models by at least 45\%. Smaller and cheaper VLMs like Gemini-2.5-Flash are still strong, falling short of the best model by only 9\%, while the open-source Qwen2.5-VL-7b \cite{qwen2_5_vl} scores 38\% less. Additionally, we find that among finetuned baselines, many degrade when generalizing to other benchmark datasets. Finally, we show that our VLM scaffold approach transfers to other image analysis tasks --  they greatly enhance zero-shot VLM performance on artifact detection for object morphology, animal anatomy, irrational interactions, and general AI-generated content (AIGC).

Our contributions are:
\begin{itemize}
    \item We propose ArtifactLens, a VLM scaffold that achieves state-of-the-art artifact detection using only hundreds of labeled examples -- orders of magnitude less than finetuning approaches.
    \item On five human artifact benchmarks (the first such multi-benchmark evaluation), our methods generalize to object morphology, animal anatomy, entity interactions, and AIGC detection.
    \item We contribute new methods for black-box optimization: `counterfactual demonstrations' for superior ICL, and `full spectrum prompting' for overcoming biases in text instruction optimization.
\end{itemize}

\section{Related Work}
\label{sec:related_work}
\subsection{Artifacts in AI-generated images}
\textbf{Human artifact detection:} Human anatomy errors in AI-generated image are a major problem animating research. Multiple works release benchmarks and models for detecting human anatomy artifacts, either with image-level labels \cite{cao2024synartifact, wang2025magicmirror} or box level predictions \cite{fang2024humanrefiner, wang2024detecting, wang2025diffdoctor}, and one new benchmark extends to video \cite{kang2025geneva}. These papers also release methods, finetuning VLMs, object detectors or segmentation models with large in-distribution datasets. Some prior works motivate fine-tuning with experiments showing that frozen VLMs perform poorly. To the contrary, we find that frozen VLMs -- properly adapted with scaffolding techniques -- performs comparably or better than fine-tuned methods with data efficiency. Our work is the first to combine these benchmarks and tests all methods, which enables testing generalization and sets a clear target for future work. 

\textbf{Artifacts in related tasks:} Aside from detection, some methods repair artifacts \cite{lu2024handrefiner, gandikota2024concept, fang2024humanrefiner, wang2024detecting, wang2025rhands}. A related task is AI-generated content (AIGC) detection; while some examples use human artifacts as a signal, the task is to detect AIGC even if there are no artifacts \cite{wang2020cnn, zhou2025aigi,tan2024rethinking, luo2024lare, li2025fakebench, zhu2023genimage}. A related and major research area is human preference modeling, which records a score for an image or image-pair \cite{xu2023imagereward, wu2023human}; while artifacts do impact scores, they are not explicitly modeled and the derived reward models have not solved the human artifact problem.
Finally, image quality assessment works increasingly use VLMs, but the target task is not human artifacts \cite{you2024depicting, wu2023q, wu2025visualquality, li2025q}.

\subsection{Optimizing black-box VLMs}
\textbf{Multi-component systems:}
With the development of capable and general LLMs (and VLMs), many research communities are exploring how to combine multiple model calls and external components \cite{yao2022react, du2023improving, chen2022program, wu2024autogen, khattab2022demonstrate}. Most relevant are systems with a fixed topology (in contrast to more autonomous systems where LLMs dictate the control flow) \cite{compound-ai-blog, anthropic2024effectiveagents}. 
Artifact detection is well-suited to this design: the task is decomposable into multiple single model-calls; non-finetuned models are strong; and single components can be optimized independently. Different communities use different terminology.
% Multi-component systems are popular in multiple communities that use different terminology. 
\textit{Compound systems} \cite{compound-ai-blog, wu2025optimas} is the term  advanced by the DSPy framework \cite{khattab2024dspy, khattab2022demonstrate}, which has most adoption in NLP, information retrieval, and in industry. We implement \SystemName\ in DSPy. Another term popular in industry is \textit{workflows}
\cite{anthropic2024effectiveagents}, often referring to structured chains of LLM calls. Still another framing is `multi-agent systems' \cite{wu2024autogen, hong2023metagpt} though it is more often applied to autonomous systems (no fixed topology).
One notable example of a compound system in computer vision is visual chain-of-thought which uses one model to crop images that are passed to VLMs  \cite{wu2024v, shao2024visual, liu2024chain}, enabling analysis in small regions. 
% We leverage this idea in \SystemName.
Our methods use this idea.

\textbf{Multimodal in-context learning (MM-ICL):} In-context learning techniques prompt LLMs with input-output demonstrations to define or improve performance on the task \cite{brown2020language}. Multimodal ICL extends this to image inputs \cite{alayrac2022flamingo}. While much work focuses on training strategies to improve performance \cite{doveh2024towards, huang2024multimodal}, others show that strong generalist VLMs are effective \cite{sun2024generative, jiang2024many}. MM-ICL is appealing because it leverages strong pretrained models and is data efficient, and we find it enormously improves artifact detection. Research in training-free ICL focuses on how to choose the best demonstrations. We consider both task-level example selection, and query-dependent example selection (or retrieval) \cite{qin2024factors, doveh2024towards}

\textbf{Text-prompt optimization} LLM and VLM performance is sensitive to the text prompt, and a well-designed instruction can clarify task ambiguity and cover edge cases \cite{schulhoff2024prompt}. The literature shows that LLMs themselves can construct good prompts \cite{yang2023large, zhou2022large}, and they can even rewrite prompts by analyzing how they performed against a development dataset \cite{khattab2024dspy, agrawal2025gepa, xiang2025self}. A few works apply LLM prompt writers for vision-language tasks, though most are for CLIP models \cite{mirza2024glov, du2024ipo, choi2025multimodal, wu2025hierarchical}.

    \begin{figure}[t]
    \centering
    \includegraphics[page=3, width=\columnwidth,clip,trim=0 775 1285 0]{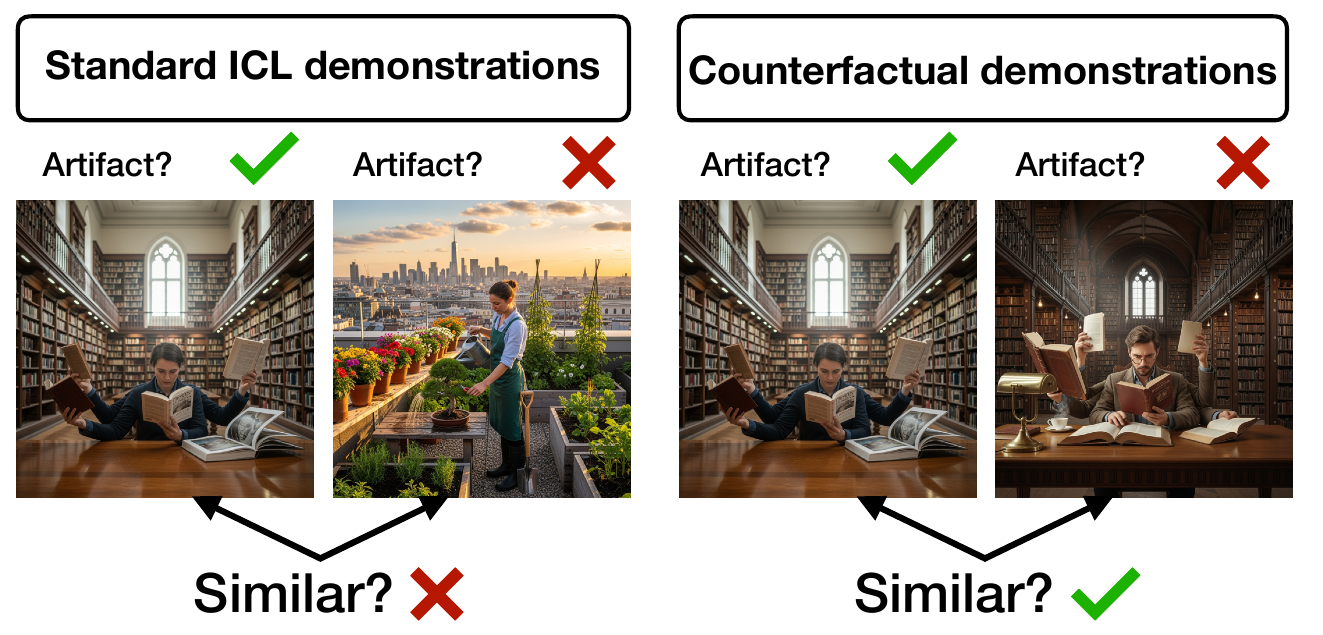}
    \captionof{figure}{
    \textbf{Counterfactual demonstrations} for in-context learning (ICL) (\cref{sec:results-icl}): typical ICL methods may not consider the relationship between different demonstrations.
    We choose demonstrations in pairs where the images are semantically similar, but with opposite artifact label -- this more clearly defines the learning task.
    }
    \label{fig:counterfactual-demos}
\end{figure}

    \begin{figure}[h]
    \centering
    \includegraphics[page=4, width=\columnwidth,clip,trim=0 755 1360 0]{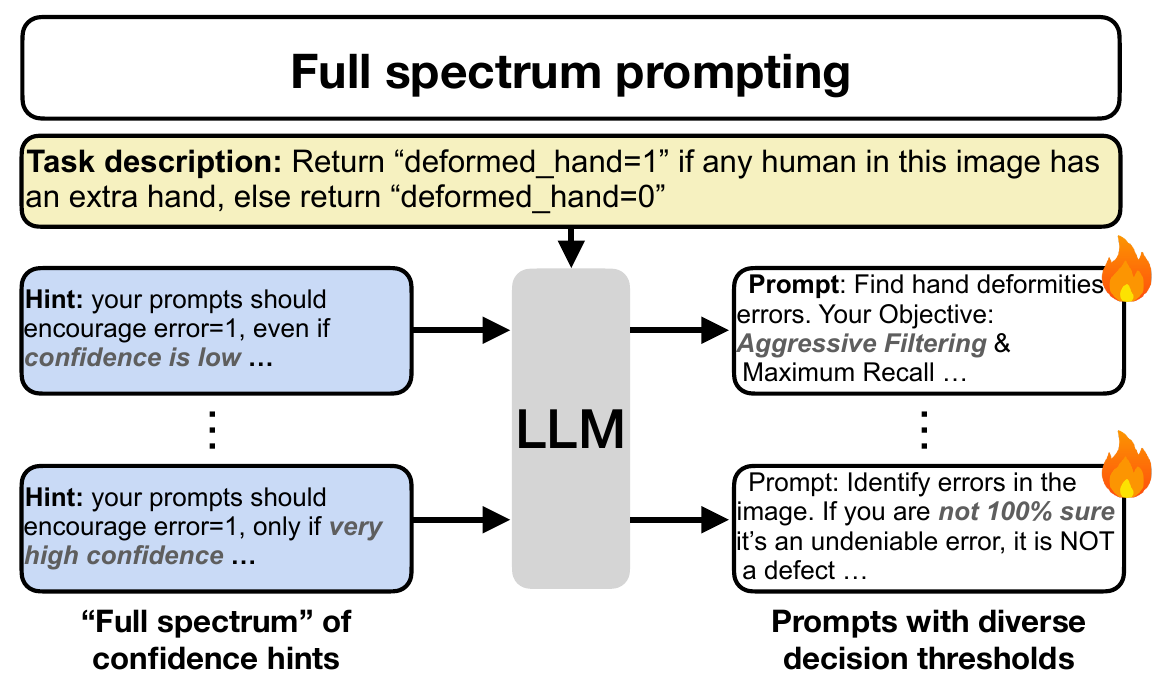}
    \captionof{figure}{
    In text optimization, LLMs map a task description (yellow) to candidate text prompts (clear boxes).
    We add hints (blue) that cover the `full spectrum' of confidence thresholds. Without this, most generated prompts are cautious -- instructing the VLM to only flag errors if confident -- which leads to worse artifact detection performance.
    }
    \label{fig:spectrum-prompting}
\end{figure}
\section{Methods}
\label{sec:methods}
Our method, \SystemName, is an artifact detector built by scaffolding pretrained VLMs. Here, we first formulate the task. Then we detail the methods: the multi-component architecture, in-context learning techniques, and text optimization techniques. All uses of the term `VLM' refer to LLM-based vision-language models \cite{alayrac2022flamingo}. 

\subsection{Task formulation}
\label{sec:methods_task_formulation}
We need a function, $f$ that classifies images, $\mathcal{I}$ into `artifact' or `not artifact', $f:\mathcal{I}\to\{1,0\}$. This formulation best aligns with the practical goals of artifact detectors because image-level binary labels are sufficient for the main use-cases of data filtering, rejection sampling, and reward modeling. Image-to-label functions are also a natural approach for leveraging vision-language models which map images to text. We use the word `detection' (rather than `classification') because we are identifying aberrations from `normal' or in-distribution behavior, similar to how `anomaly detection' can use image-level labels \cite{bergmann2019mvtec, hendrycks2018deep}.

Why not some other formulation? One alternative defines $K$ artifact sub-classes for multi-label binary prediction: $f:\mathcal{I}\to\{1,0\}^K$, which is appealing because all the training datasets define a taxonomy of subclasses. We prefer the single-label formulation because multilabel classification is not necessary for the downstream applications and because different benchmarks use different sub-label taxonomies so comparison is difficult. Still, our method can support this format if future research wishes to explore it. Another formulation is bounding-box detection \cite{fang2024humanrefiner, wang2025generated, wang2024detecting}, however this is a more challenging task to learn, requires more annotation effort, and is not necessary for most applications.

\subsection{Multi-component architecture}
\label{sec:methods-architecture}
Our system, \SystemName\ has a multi-component architecture in a fixed topology shown in \cref{fig:methods} -- this is often called a compound system or workflow \cite{compound-ai-blog, anthropic2024effectiveagents}. There are $K$ specialists, that each use VLMs to predict one type of error such as `deformed face' or `distorted hand' -- the choices of error type is determined by the dataset's labeling taxonomy. Each specialist predicts for its own type, and those predictions are  aggregated into a single label, $y$. Intuitively, it should be easier for VLMs to identify one specific error type compared to multiple types, especially if we can optimize the `missing arm' specialist (for example) with annotations for `missing arm'.

\begin{table*}[t]
\centering
\caption{
Benchmark suite: five benchmarks and their attributes. More details are in 
\cref{sec:result_benchmarks}.
}
\label{table:benchmark-attributes}
% \resizebox{\columnwidth}{!}{
\scriptsize
\begin{tabular}{@{}lllllccc@{}}
\toprule
 \textbf{Benchmark} & \textbf{\begin{tabular}[c]{@{}l@{}}Release\\ date\end{tabular}} & \textbf{\begin{tabular}[c]{@{}l@{}}Label \\ type\end{tabular}} & \textbf{T2I models} & \textbf{\begin{tabular}[c]{@{}l@{}}T2I prompt \\ sources\end{tabular}} & \multicolumn{1}{l}{\textbf{\begin{tabular}[c]{@{}l@{}}\# sublabel \\ types\end{tabular}}} & \multicolumn{1}{l}{\textbf{\begin{tabular}[c]{@{}l@{}}\# samples\\ train/test\end{tabular}}} & \multicolumn{1}{l}{\textbf{\begin{tabular}[c]{@{}l@{}}Artifact \\ rate\end{tabular}}} \\ \midrule
SynArtifact \cite{cao2024synartifact} & Feb '24 & Box & SD1.0-2.1, DALL-E 3 & \begin{tabular}[c]{@{}l@{}}ImageNet, COCO, \\ Midjourney users\end{tabular} & 13 & 53 / 266 & 65.4\% \\
AbHuman \cite{fang2024humanrefiner} & Jul '24 & Box & SDXL & \begin{tabular}[c]{@{}l@{}}LAION‑5B, Human‑\\ Art, GPT‑3.5\end{tabular} & 7 & 45k / 11k & 70.2\% \\
HAD \cite{wang2024detecting} & Nov '24 & Box & \begin{tabular}[c]{@{}l@{}}DALLE‑2, DALLE‑3, \\ SDXL, Midjourney\end{tabular} & GPT‑4 generated & 14 & 33k / 4k & 70.6\% \\
AIGC-HA \cite{wang2025generated} & Nov '24 & Box & VidProM (PIKA split) & \begin{tabular}[c]{@{}l@{}}Inherited from \\ VidProM PIKA\end{tabular} & 40 & 42K / 1K & 65.4\% \\
MagicBench \cite{wang2025magicmirror} & Sep '25 & Image & \begin{tabular}[c]{@{}l@{}}FLUX.1‑dev/schnell, Kolors 1.0, \\ SD3/3.5, Midjourney‑v6.1, private\end{tabular} & \begin{tabular}[c]{@{}l@{}}Pick‑a‑Pic, \\ human-generated\end{tabular} & 7 & 167k / 17k & 66.6\% \\ \bottomrule
\end{tabular}
% }`
\end{table*}

% \begin{table*}[t]
% \centering
% \caption{
% Benchmark suite: five benchmarks and their attributes
% . The `\# samples' is after our adaptations described in the text.
% }
% \label{table:benchmark-attributes}
% \resizebox{\textwidth}{!}{
% \scriptsize
% \begin{tabular}{lllllccc}
% \textbf{Benchmark} & \textbf{Release date} & \textbf{Label  type} & \textbf{T2I models} & \textbf{T2I prompt  sources} & \multicolumn{1}{l}{\textbf{\# sublabel types}} & \multicolumn{1}{l}{\textbf{\# samples train/test}} & \multicolumn{1}{l}{\textbf{Artifact rate}} \\ \hline
% SynArtifact \cite{cao2024synartifact} & Feb$\sim$’24 & Image & SD1.0-2.1, DALL-E 3 & ImageNet, COCO,  Midjourney users & 13 & 53 / 266 & 65.4\% \\
% AbHuman \cite{fang2024humanrefiner} & Jul$\sim$’24 & Box & SDXL & LAION‑5B, Human‑Art, GPT‑3.5 & 18 & 45k / 11k & 70.2\% \\
% HAD \cite{wang2024detecting} & Nov$\sim$’24 & Box & DALLE‑2, DALLE‑3,  SDXL, Midjourney & GPT‑4 generated & 6 & 33k / 4k & 70.6\% \\
% AIGC-HA \cite{wang2025generated} & Nov$\sim$’24 & Box & VidProM (PIKA split) & Inherited from  VidProM PIKA & 12 & / 1000 & 65.4\% \\
% MagicBench \cite{wang2025magicmirror} & Sep$\sim$’25 & Image & \begin{tabular}[c]{@{}l@{}}FLUX.1‑dev/schnell, Kolors 1.0, \\ SD3/3.5, Midjourney‑v6.1, private\end{tabular} & Pick‑a‑Pic,  human-generated & 7 & 167k / 17k & 66.6\%
% \end{tabular}
% }
% \end{table*}

As in \cref{fig:methods}, \SystemName\ takes an image and routes a copy to each specialist. Consider the specialist for `extra arm'. We crop the image into candidate regions of interest -- in this case around each human, which are identified using GoundingDino for object detection \cite{liu2024grounding}. (Different specialists have different detection targets, \cref{sec:appendix-methods}). This is because artifacts can be small and VLMs perform better if processing the zoomed-in region-of-interest \cite{shao2024visual, liu2024chain, wu2024v}. Each crop is passed to the VLM, along with the original image, and prompted for a binary prediction for the artifact, $y_{crop}\in\{1,0\}$. If any crop has $y_{crop}=1$ then the specialist prediction is $y_{spec}=1$, otherwise $y_{spec}=0$. More formally, we take the logical-or over the $L$ crops: $y_{spec}=\bigvee_{\ell=1}^{L} y_{crop}^\ell$. Then, the $K$ specialist predictions are aggregated in the same way: $y=1$ if any specialist has $y_{spec}=1$, otherwise $y=0$ and again, this is a logical or: $y=\bigvee_{k=1}^{K} y_{spec}^k$. The system is implemented as a DSPy program \cite{khattab2024dspy, khattab2022demonstrate}. 

But how are the specialist VLMs prompted? A key idea in multi-component optimization \cite{khattab2024dspy, wu2024autogen, schulhoff2024prompt} is that rather than manual prompt engineering, we start with a seed prompt (\textit{signature} in DSPy) that is clear and concise -- for example ``return artifact=1 if there is a human with a missing foot'' -- then we apply automatic methods to improve the prompt. In DSPy terminology \cite{khattab2024dspy}, we apply \textit{optimizers} that modify a DSPy \textit{program}. Optimizing each of the $K$ specialists requires a training dataset, $(x^k,y^k)\sim\mathcal{D}^k_t$ for sublabel $k$ (e.g. `deformed face').  We set $y^k=1$ if that sublabel is 1, but only set $y^k=0$ if all sublabels are 0; this filters images with other error types. We can also sample a validation set, $\mathcal{D}^k_v$ from the train set.  The next sections detail the optimization techniques: in-context learning and prompt optimization, which we apply to specialists independently.

\begin{table*}[ht]
\centering
\caption{Human artifact detection results on our benchmarks suite (see \cref{sec:result_benchmarks}). 
The \emph{Finetuned baselines} are VLMs or object detectors trained on a single artifact dataset; an asterisk (\(*\)) marks in-distribution training for that benchmark. 
\emph{VLMs (zero-shot)} are prompted with a list of target artifacts. 
\SystemName\ rows use the same backbones and show gains (green deltas) over their zero-shot counterparts.}
\label{tab:main-results}
\setlength{\tabcolsep}{2pt}
\renewcommand{\arraystretch}{1.15}

\resizebox{\textwidth}{!}{%
\begin{tabular}{
  @{}
  l
  >{\hspace{20pt}}l
  |
  S@{\hspace{4pt}}l        % Overall F1
  S@{\hspace{4pt}}l        % Overall Precision
  S@{\hspace{4pt}}l        % Overall Recall (extra pad to the border)
  |
  S@{\hspace{4pt}}l        % AIGC-HA
  S@{\hspace{4pt}}l        % HAD
  S@{\hspace{4pt}}l        % AbHuman
  S@{\hspace{4pt}}l        % MagicBench
  S@{\hspace{4pt}}l        % SynArtifact
  @{}
}

% ---------- Header ----------
\rowcolor{HeaderBG}
 &  &
   \multicolumn{6}{c|}{\textbf{Overall}} &
   \multicolumn{10}{c}{\textbf{Per-benchmark F1}} \\
\rowcolor{HeaderBG}
 &  &
   \multicolumn{2}{c}{\textbf{F1}} &
   \multicolumn{2}{c}{\textbf{Precision}} &
   \multicolumn{2}{c|}{\textbf{Recall}} &
   \multicolumn{2}{c}{\textbf{AIGC-HA}~\cite{wang2025generated}} &
   \multicolumn{2}{c}{\textbf{HAD}~\cite{wang2024detecting}} &
   \multicolumn{2}{c}{\textbf{AbHuman}~\cite{fang2024humanrefiner}} &
   \multicolumn{2}{c}{\textbf{MagicBench}~\cite{wang2025magicmirror}} &
   \multicolumn{2}{c}{\textbf{SynArtifact}~\cite{cao2024synartifact}} \\
\midrule

% ---------- Finetuned baselines ----------
\multirow{4}{*}{\begin{tabular}[c]{@{}l@{}}\textbf{Finetuned baselines}\end{tabular}}
 & MagicAssessor~\cite{wang2025magicmirror}
   & 0.754 & &
     0.808 & &
     0.720 & &
     0.702 & &
     0.814 & &
     0.738 & &
     0.754* & &
     0.778 & \\
 & HADM~\cite{wang2024detecting}
   & 0.758 & &
     0.773 & &
     0.759 & &
     0.611 & &
     \bfseries 0.865* & &
     0.729 & &
     0.810 & &
     0.773 & \\
 & AHD ~\cite{wang2025generated}
   & 0.620 & &
     0.757 & &
     0.554 & &
     0.791* & &
     0.537 & &
     0.791 & &
     0.581 & &
     0.645 & \\
 & Diffdoctor~\cite{wang2025diffdoctor}
   & 0.544 & &
     0.783 & &
     0.425 & &
     0.372 & &
     0.593 & &
     0.503 & &
     0.636 & &
     0.618 & \\
\midrule

% ---------- VLMs (zero-shot) ----------
\multirow{5}{*}{\begin{tabular}[c]{@{}l@{}}\textbf{VLMs (zero-shot)}\end{tabular}}
 & Gemini-2.5-pro
   & 0.560 & &
     \bfseries 0.817 & &
     0.435 & &
     0.504 & &
     0.451 & &
     0.663 & &
     0.623 & &
     0.791 & \\
 & GPT-4o
   & 0.217 & &
     0.807 & &
     0.135 & &
     0.347 & &
     0.038 & &
     0.377 & &
     0.106 & &
     0.437 & \\
 & Gemini-2.5-flash
   & 0.369 & &
     0.789 & &
     0.254 & &
     0.308 & &
     0.198 & &
     0.542 & &
     0.427 & &
     0.706 & \\
 & GPT-4o-mini
   & 0.087 & &
     0.816 & &
     0.047 & &
     0.108 & &
     0.002 & &
     0.144 & &
     0.093 & &
     0.103 & \\
 & QwenVL-2.5-7B
   & 0.017 & &
     0.646 & &
     0.009 & &
     0.023 & &
     0.001 & &
     0.019 & &
     0.025 & &
     0.011 & \\
\midrule

% ---------- ArtifactLens (ours) ----------
\rowcolor{OursBG}
 & Gemini-2.5-pro
   & \bfseries 0.817 & \deltaposmed{0.257}
   & 0.755 & \deltaneg{0.062}
   & \bfseries 0.906 & \deltaposhigh{0.471}
   & 0.759 & \deltaposmed{0.255}
   & 0.840 & \deltaposmed{0.389}
   & \bfseries 0.841 & \deltapos{0.178}
   & \bfseries 0.849 & \deltapos{0.226}
   & \bfseries 0.798 & \deltapos{0.007} \\
\rowcolor{OursBG}
 & GPT-4o
   & 0.802 & \deltaposhigh{0.585}
   & 0.766 & \deltaneg{0.041}
   & 0.865 & \deltaposhigh{0.730}
   & \bfseries 0.796 & \deltaposhigh{0.449}
   & 0.814 & \deltaposhigh{0.776}
   & 0.822 & \deltaposhigh{0.445}
   & 0.817 & \deltaposhigh{0.711}
   & 0.760 & \deltaposmed{0.323} \\
\rowcolor{OursBG} \textbf{\SystemName\ (ours)}
 & Gemini-2.5-flash
   & 0.747 & \deltaposmed{0.378}
   & 0.752 & \deltaneg{0.037}
   & 0.778 & \deltaposhigh{0.524}
   & 0.640 & \deltaposmed{0.332}
   & 0.792 & \deltaposhigh{0.594}
   & 0.831 & \deltaposmed{0.289}
   & 0.795 & \deltaposmed{0.368}
   & 0.675 & \deltaneg{0.031} \\
\rowcolor{OursBG}
 & GPT-4o-mini
   & 0.571 & \deltaposhigh{0.484}
   & 0.805 & \deltaneg{0.011}
   & 0.451 & \deltaposhigh{0.404}
   & 0.462 & \deltaposmed{0.354}
   & 0.654 & \deltaposhigh{0.652}
   & 0.690 & \deltaposhigh{0.546}
   & 0.611 & \deltaposhigh{0.518}
   & 0.438 & \deltaposmed{0.335} \\
\rowcolor{OursBG}
 & QwenVL-2.5-7B
   & 0.501 & \deltaposhigh{0.484}
   & 0.780 & \deltapos{0.135}
   & 0.376 & \deltaposmed{0.367}
   & 0.368 & \deltaposmed{0.345}
   & 0.479 & \deltaposhigh{0.478}
   & 0.611 & \deltaposhigh{0.592}
   & 0.544 & \deltaposhigh{0.519}
   & 0.402 & \deltaposmed{0.391} \\
   
\bottomrule
\end{tabular}%
}
\end{table*}

\subsection{In-context learning methods}
\label{sec:results-icl}
In-context learning (ICL) improves the prompt by  putting example input-output pairs, $(x,y)$ in the VLM prompt \cite{alayrac2022flamingo, sun2024generative}. The key challenge is choosing the best demonstrations from the optimization training set. We fix $m=10$ demonstrations.

The first class of methods involve random sampling. The simplest (called \texttt{LabeledFewShot} (LFS) in DSPy \cite{khattab2022demonstrate}) takes $m$ random demonstrations \cite{alayrac2022flamingo}, and results will show this does improve over zero-shot baselines. We propose a simple extension, \texttt{LabeledFewShotWRandomSearch} (LFSRS) which samples multiple random demonstration sets, evaluates the VLM against the validation dataset, choosing the one with the highest score. 

The second class of methods retrieve demonstrations conditioned on the query image \cite{yang2023re, doveh2024towards}; the training set serves as the retrieval corpus.. We implement \texttt{DynamicFewShot}, which retrieves the nearest images in CLIP-space \cite{radford2021learning}, with the intuition that semantically similar images make better demonstrations. We enforce that the $m$ examples are class-balanced. Our main results use \texttt{DynamicFewShot}, since it gave the best scores.

Extending these ideas, we propose \textit{counterfactual prompting} for binary ICL tasks (\cref{fig:counterfactual-demos}). We show ICL examples in pairs: one with artifacts and one without that are otherwise visually very similar. Intuitively, this better isolates the task definition -- if the main difference between two images is the presence of an artifact, then the task should be easier to learn. This idea can be applied to both approaches -- random sampling and retrieval. In our implementation, we use CLIP distance to define the similarity.

\subsection{Text-prompt optimization methods}
\label{sec:methods-text_optimization}
VLM performance depends on the text instructions that, together with ICL samples, define the task \cite{schulhoff2024prompt}. Automatic prompt optimization searches over candidate prompts given a validation dataset. A strong baseline is COPRO \cite{khattab2024dspy}, a DSPy extension of OPRO (Optimization by PROmpting, \cite{yang2023large}. It takes the original prompt, asks an LLM to generate a pool of candidate instructions, and evaluates them against the validation set on the target task. Then it iteratively picks the best-performing instructions, passes them back to the instruction generator LLM with the scores to generate a new candidate pool, and evaluates again. We apply ICL and text optimization together, so prompts include images.

We propose a new technique for instruction generation: \textit{full-spectrum prompting} (\cref{fig:spectrum-prompting}). We observed that LLM-generated instructions tend to be cautious, often instructing the VLM to ``only label artifacts if you are very confident'' (\cref{sec:supp-fullspectrum-prompting}). However, we find better performance with the opposite guidance, such as ``apply the artifact label even if you have low confidence'' -- we hypothesize that VLMs are biased against the artifact label \cite{vo2025vision}. This technique integrates easily with existing methods like COPRO. We modify the instruction generator prompt to create instructions corresponding to an X\% confidence threshold, where $X\sim\mathcal{U}[0,100]$ (\cref{sec:supp-fullspectrum-prompting}). This ensures the candidate prompt pool covers the `full spectrum' of decision thresholds. (Note that simply instructing the LLM to ``be diverse'' in confidence thresholds proved ineffective.) In all experiments, we use the same LLM/VLM for both instruction generation and evaluation.  Additionally, an even simpler approach is competitive: generate the prompt pool by simply adding the confidence hints directly to the seed instruction, skipping the LLM.

\section{Results}
\label{sec:results}
The main results are for image-level artifact detection as formulated in \cref{sec:methods_task_formulation}. 
We combine existing benchmarks into one test suite in \cref{sec:result_benchmarks}. We report the main results in  \cref{sec:results_main_results}, and then ablate the components of our \SystemName\ approach in \cref{sec:ablations}. Finally we show that black box optimization of pretrained VLMs can benefit other image atribute tasks in \cref{sec:results_other_detection_tasks}

\subsection{Benchmark suite \& baselines}
\label{sec:result_benchmarks}
\textbf{Benchmarks:} 
There are many prior works on artifact detection, but they only evaluate on their own test sets. We combine them into a single benchmark suite, which allows  studying cross-dataset generalization. They are: SynArtifact \cite{cao2024synartifact}, AbHuman \cite{fang2024humanrefiner}, Human Artifact Dataset (HAD) \cite{wang2024detecting}, AIGC Human-Aware-1K \cite{wang2025generated}, and MagicBench \cite{wang2025magicmirror}. (DiffDoctor \cite{wang2025diffdoctor}  released a model but not a dataset). 

\begin{table*}[t]
\centering
\caption{
Ablations of VLM optimization components of scaffolding, in-context learning (ICL), and text optimization (TextOpt). The metric is mean positive-class F1 on MagicBench \cite{wang2025magicmirror}.
}
\label{table:ablations-main}
\setlength{\tabcolsep}{10pt}
\renewcommand{\arraystretch}{1.15}
\resizebox{\textwidth}{!}{
\begin{tabular}{l|cccccc}
\rowcolor{HeaderBG}
 & Gemini-2.5-pro & GPT-4o & Gemini-2.5-flash & GPT-4o-mini & Qwen2.5-VL7b & Average \\
\midrule
\rowcolor{OursBG}
ArtifactLens & 0.849 & 0.817 & 0.795 & 0.611 & 0.544 & 0.780 \\
\midrule
\textbf{With specialists (multi-component architecture)} & \multicolumn{6}{l}{} \\
w/o In-Context-Learning
  & 0.800 \deltaneg{0.043}
  & 0.358 \deltaneg{0.460}
  & 0.756 \deltaneg{0.050}
  & 0.147 \deltaneg{0.505}
  & 0.125 \deltaneg{0.419}
  & 0.515 \deltaneg{0.265} \\
w/o Text optimization
  & 0.846 \deltapos{0.003}
  & 0.819 \deltapos{0.001}
  & 0.733 \deltaneg{0.073}
  & 0.633 \deltaneg{0.019}
  & 0.647 \deltaposmed{0.103}
  & 0.758 \deltaneg{0.022} \\
w/o In-Context-Learning, w/o Text Optimization
  & 0.733 \deltaneg{0.110}
  & 0.271 \deltaneg{0.547}
  & 0.527 \deltaneg{0.279}
  & 0.165 \deltaneg{0.487}
  & 0.154 \deltaneg{0.390}
  & 0.424 \deltaneg{0.356} \\
\midrule
\textbf{W/o specialists (single VLM)} & \multicolumn{6}{l}{} \\
ArtifactLens
  & 0.834 \deltaneg{0.009}
  & 0.742 \deltaneg{0.076}
  & 0.723 \deltaneg{0.083}
  & 0.615 \deltaneg{0.037}
  & 0.549 \deltapos{0.005}
  & 0.729 \deltaneg{0.051} \\
w/o In-Context-Learning
  & 0.767 \deltaneg{0.076}
  & 0.252 \deltaneg{0.566}
  & 0.712 \deltaneg{0.094}
  & 0.108 \deltaneg{0.544}
  & 0.100 \deltaneg{0.444}
  & 0.460 \deltaneg{0.320} \\
w/o Text optimization
  & 0.823 \deltaneg{0.020}
  & 0.724 \deltaneg{0.094}
  & 0.558 \deltaneg{0.248}
  & 0.632 \deltaneg{0.020}
  & 0.449 \deltaneg{0.095}
  & 0.684 \deltaneg{0.096} \\
w/o In-Context-Learning, w/o Text Optimization
  & 0.579 \deltaneg{0.264}
  & 0.173 \deltaneg{0.645}
  & 0.389 \deltaneg{0.417}
  & 0.080 \deltaneg{0.572}
  & 0.070 \deltaneg{0.474}
  & 0.305 \deltaneg{0.475} \\
\bottomrule
\end{tabular}
}
\end{table*}

The benchmark dataset attributes are in \cref{table:benchmark-attributes} (extended in \cref{sec:appendix-benchmark_details}). Our benchmarks are consistent with our per-image label formulation, while the others use bounding-box detection metrics, which we convert to image-level labels. Two benchmarks -- SynAritfact and MagicBench -- also label non-human artifacts which we ignore; similarly, AbHuman has a `not human' error class which we remove. For all three, we filter images not containing humans using a simple VLM query with GPT-4o \cite{achiam2023gpt}. We sample 2500 images for each benchmark's test set, or the entire test set for smaller benchmarks, namely AIGC Human-Aware-1K (1K), and SynArtifact (266).
For SynArtifact we use the train set for testing and vice versa because the original test set is small (53 images) after human filtering.

All benchmarks have artifact prevalence between 50\% and 70\%. The choice of text-to-image model varies, inducing a natural distribution shift: for example SynArtifact is the earliest and includes Dalle-2 \cite{ramesh2022hierarchical}, while MagicMirror is the most recent and includes FLUX.1-dev/schnell \cite{blackforestlabs2024flux}. They target similar error types -- namely missing, extra, or deformed body parts, though their exact taxonomies differ. However since they were labeled by different groups with different labeling guidelines, this is another source of distribution shift \cite{recht2019imagenet}. \Cref{sec:appendix-benchmark_details} shows label taxonomies and label distributions -- `hand defects' is the most prevalent class.

\textbf{Metrics:} The primary goal is detecting artifacts, so the main metrics are positive class F1, precision, and recall; this is in line with prior works \cite{cao2024synartifact, wang2025magicmirror}. We show all three averaged over the benchmarks, plus F1 for each benchmark.

\textbf{Baselines models:} 
The baselines are models finetuned on artifact datasets. MagicAssessor tunes a VLM on MagicMirror, the train set of MagicBench \cite{wang2025magicmirror}; we ignore non-human artifacts. Two are object detectors: HADM is trained on HAD \cite{wang2024detecting}, while AHD is  trained on a synthetic dataset different to its benchmark, AIGC-HA \cite{wang2025generated}. DiffDoctor \cite{wang2025diffdoctor} has a pixelwise prediction model. For the detection and pixel models, we use the author's recommended thresholds and map boxes to image-level labels. To the best of our knowledge, these are the top artifact detection models. The authors of SynArtifact \cite{cao2024synartifact} and AbHuman \cite{fang2024humanrefiner} did not release their detection models.

\begin{figure}[hb]
    \centering

    {%
    \captionsetup{type=table}%
    \caption{
        Performance of \SystemName\ on other (non-human-artifact) tasks, showing the generality of the methods (\cref{sec:results_other_detection_tasks}).
    }
    \label{tab:other-tasks}
    }

    {
    \setlength{\tabcolsep}{2pt}
    \renewcommand{\arraystretch}{1.15}

    \resizebox{\columnwidth}{!}{%
    \begin{tabular}{@{}l|
        S[table-format=1.3]
        S[table-format=1.3]| 
        S[table-format=1.3]
        S[table-format=1.3]@{}}

    \rowcolor{HeaderBG}
      & \multicolumn{2}{c|}{\textbf{Zero-shot VLMs}}
      & \multicolumn{2}{c}{\textbf{Optimized VLMs (ours)}} \\

    \rowcolor{HeaderBG}
    \textbf{Detection target}
      & \textbf{Gemini-2.5-Pro}
      & \textbf{GPT-4o}
      & \textbf{Gemini-2.5-Pro}
      & \textbf{GPT-4o} \\
    \midrule

    AI-generated image detection
      & 0.922 & 0.896
      & {\cellcolor{OursBG}\diffpos{0.998}{0.002}}
      & {\cellcolor{OursBG}\diffpos{0.998}{0.134}} \\

    Irrational object interaction
      & 0.456 & 0.229
      & {\cellcolor{OursBG}\diffpos{0.531}{0.188}}
      & {\cellcolor{OursBG}\diffpos{0.454}{0.260}} \\

    Animal artifact
      & 0.488 & 0.352
      & {\cellcolor{OursBG}\diffpos{0.676}{0.075}}
      & {\cellcolor{OursBG}\diffpos{0.612}{0.225}} \\

    Object morphology artifact
      & 0.401 & 0.230
      & {\cellcolor{OursBG}\diffpos{0.403}{0.076}}
      & {\cellcolor{OursBG}\diffpos{0.364}{0.102}} \\

    \bottomrule
    \end{tabular}%
    }
    }

    \vspace{1em}

    \includegraphics[
        page=5,
        width=\columnwidth,
        clip,
        trim=0 580 620 0
    ]{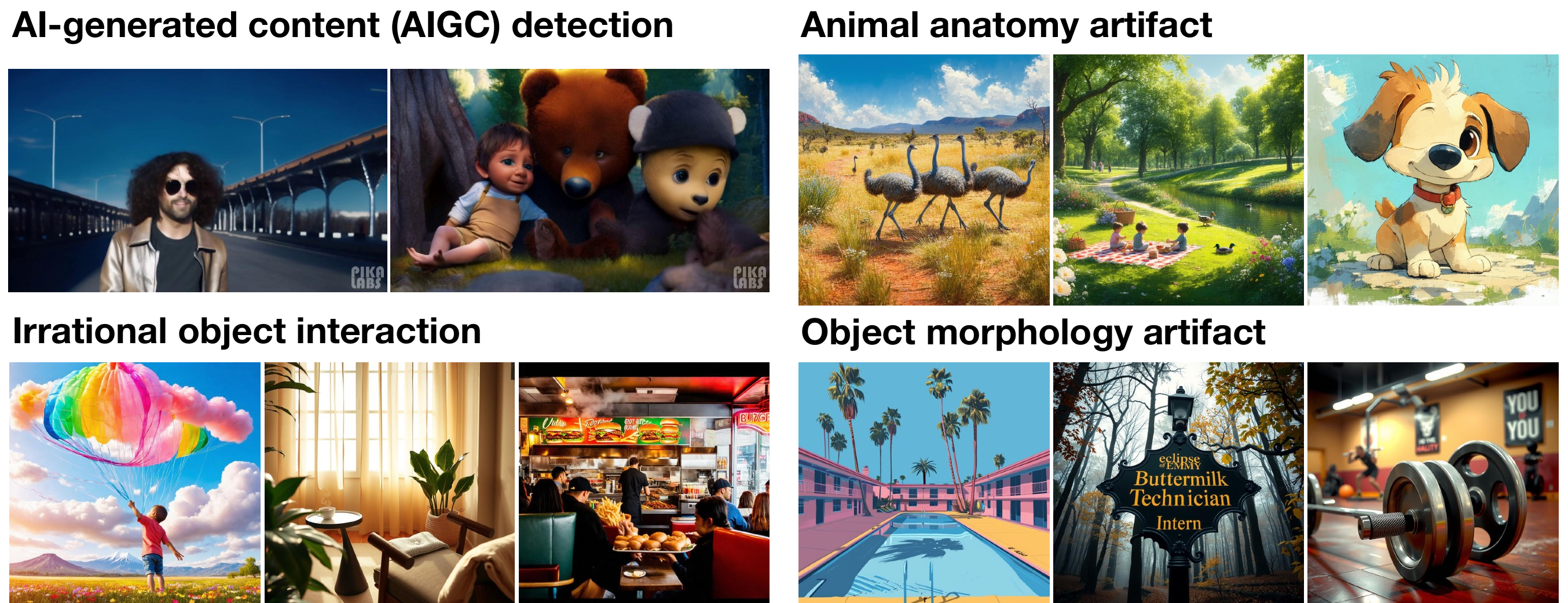}

    \caption{
        Examples for the other (non-human-artifact) tasks where \SystemName\ performs well (\cref{sec:results_other_detection_tasks}).
    }
    \label{fig:other-tasks}

\end{figure}

The other important baselines are zero-shot pretrained VLMs. We include: two top-performing models, Gemini-2.5-Pro \cite{gemini25pro} and GPT-4o \cite{gpt4o} (GPT-5 \cite{gpt5} scores are close to GPT-4o, so we skip it in main tables for brevity); two smaller models since expense is a consideration, namely Gemini-2.5-flash and GPT-4o-mini; and the open-source QwenVL-2.5-7B \cite{qwen2_5_vl}. The text prompt is: ``In this AI generated image, return artifact=1 if you see any of these human artifacts: \{artifacts description\}'' (\cref{sec:appendix-results}).

\textbf{\SystemName:} Our system is instantiated with the same VLMs as the zero-shot baselines. We use a training set of up to 5,000 and from that, sample 500 as a validation set. 
Note however that our results are still very competitive with much much less data, as we show in \cref{sec:ablations}.

\subsection{Results on the benchmark suite}
\label{sec:results_main_results}
\textbf{Main results}
\Cref{tab:main-results} shows that our \SystemName\ with Gemini-2.5-Pro, has the strongest overall F1  and recall. With GPT-4o, \SystemName\ scores only 2\% lower, and the smaller (and cheaper) Gemini-2.5-Flash is only 9\% lower. GPT-4o-mini and the open-source QwenVL-7B are competitive but weaker. 

Compare \SystemName\ with the zero-shot VLMs: our simple optimization increases the VLM's F1 score by 31\% for Gemini-2.5-pro, and by more than 54\% for all others. (The poor zero-shot scores are driven by very low recall -- they rarely classify artifacts.) This supports our earlier framing: ``VLMs can already detect artifacts'' -- they just require data-efficient \textit{adaptation} (or alternatively \textit{alignment} or \textit{elicitation}) \cite{zhou2023lima, michaud2024elicitation}. 
To extend this idea: observe that GPT-4o is significantly worse than Gemini-2.5-Pro in zero-shot (by 61\% F1), however it catches up after optimization. One could say that GPT-4o has equally-good artifact detection \textit{capability}, but is more poorly aligned to the task. This worse alignment is possibly a language bias against the `artifact' class \cite{vo2025vision}.

% \JB{could say that low zero-shot scores are consistent with prior work, and that SOME of them used this to motivate large data collection efforts \cite{wang2025diffdoctor, wang2025magicmirror}}

Among the `Finetuned Baselines' in \cref{tab:main-results}, MagicAssessor and HADM \cite{wang2024detecting} are the strongest, falling short of \SystemName\ by 7\% in overall F1. HAD performs well on its in-distribution benchmark (indicated by a `*'), generalizes well to MagicBench and SynArtifact, but drops off for AIGC-HA and AbHuman.  MagicAssessor is strong on multiple benchmarks, but notably underperforms on its own benchmark (MagicBench) compared to HAD. One possible explanation is that its training data has a broader error taxonomy including object morphology and animal artifacts, and this makes learning human artifacts harder. If true, this is an argument against large-scale fine-tuning for artifacts -- one could instead apply \SystemName\ for different artifact types independently. AHD performs well on the in-distribution AIGC-HA and on AbHuman, but struggles to generalize to others. DiffDoctor performs the worst, likely due to its more limited taxonomy of artifact errors that focuses on missing or extra body parts and less on deformities.

\textbf{Human baseline}
To understand the subjectivity of human artifact detection, we perform a human evaluation for the main sub-label of `hand artifact detection'. Ten subjects each reviewed 600 images: 200 from each of MagicBench, HAD, and AbHuman. The `majority vote' prediction scores 0.701 F1 with 0.809 precision and 0.618 recall. Their F1 score slightly underperforms our top models, which is driven by much lower recall, though slightly higher precision. For inter-rater agreement, the pairwise Cohen’s $\kappa$ is $0.639 \pm 0.078$ \cite{cohen1960coefficient}, which is between `moderate' and `substantial' agreement according to  \citet{landis1977measurement}.

These results show that \SystemName\ models perform well compared to humans, and that the artifact detection is a moderately subjective task. In \cref{sec:appendix-results} we show sample images with high disagreement and discuss likely causes. These include: small artifact region; blur or dark regions; stylized or artistic scenes; complex scenes with many hands; abnormal hand size; and partial occlusions.

\subsection{Ablations}
\label{sec:ablations}

\textbf{Ablating the major components:}
\Cref{table:ablations-main} ablates the main strategies of \SystemName, namely multi-component architecture (we will call it `specialists'), in-context learning (ICL), and text prompt optimization (TextOpt). The metric is F1 averaged over the benchmark suite, and we show results per model, and averaged over all models. 

Ablating the specialists (while still doing ICL and \mbox{TextOpt}) reduces F1 by 0.05 on average. Therefore, practitioners can make a tradeoff: provided they can tolerate performance loss, they can switch to single VLM calls with optimization, thus saving on inference cost from multiple VLM calls. It is also interesting to study the scaffold ablation in the `w/o ICL, w/o TextOpt' case; this  shows that zero-shot prompting with specialists has more significant benefit ($+0.110$ F1). Both with and without specialists, ablating ICL (resp. $-0.265$ and $-0.32$) is more significant that \mbox{TextOpt} (resp. $-0.022$ and $0.096$).

\textbf{Ablating optimization strategies:} 
We proposed two new techniques for black-box optimization, which we ablate against baseline techniques in \cref{table:ablations-optim-methods}. We start with a zero-shot VLM and apply one single technique. 

For ICL, ablating our counterfactual demonstrations on \texttt{DynamicFewShot} (`w/o cf'), the F1 score drops by $0.036$. Retrieval also beats random sampling, \texttt{LabeledFewShot}. In text optimization, ablating the confidence prompting hint from COPRO reduces F1 by $0.049$. 
% \JB{MIPRO?}

\textbf{Ablation of dataset size} 
Although our main results use 5,000 training samples, strong results are possible with much smaller datasets. This is significant for ensuring easy deployments in practical settings. Our experiment scales down the train set to 400 and samples 200 for the validation set. In this setting, \SystemName\ achieves F1 over the benchmark suite of 0.744, only 9\% less than our strongest model. 
The gains are due to ICL having a larger candidate pool and from a larger validation set  for test accuracy.

\subsection{Results on other attribute detection tasks}
\label{sec:results_other_detection_tasks}
Although we have focused on detecting human artifacts, most of the ideas presented here are not specific to that task. We hypothesize that many other vision tasks would benefit from adapting pretrained VLMs using scaffolding, in-context learning (ICL), or text optimization (TextOpt). We therefore apply ICL and TextOpt to several image analysis tasks.  The first is AI-generated content detection from AIGI-Holmes \cite{zhou2025aigi}; this can use semantic or low level cues to detect AIGC. The second set of tasks are (non-human) artifact detection from MagicBench \cite{wang2025magicmirror}, specifically animal artifacts, irrational object interactions, and object morphology.  \Cref{tab:other-tasks} shows that our optimization greatly improves VLMs compared to the zero-shot case (though we do not compare against in-domain methods that have many more features). We find that simple optimization considerably improves VLM baselines.

% https://chatgpt.com/c/68fbdda5-b51c-8329-bc6f-39e0aac5e212 

\FloatBarrier

\section{Conclusions}
\label{sec:conclusions}
We have shown that pretrained VLMs are strong artifact detectors, if provided the right scaffolding. The clearest impact is on the creative image generation field -- we hope future papers leverage \SystemName\ to improve generative models, and others further explore VLM scaffolding for artifact detection.

More generally however, we hope to inspire more vision-language researchers to explore VLM scaffolding. NLP and Information Retrieval have vibrant communities researching compound systems, in-context learning, and prompt optimization -- they enable new capabilities with great data efficiency by leveraging strong foundation models. Computer vision researchers should also pursue these direction. There is enormous opportunity to improve existing VLM adaptation methods specifically for vision and vision-language tasks,  and opportunity to apply them to more applications.
% --- Impact Statement (required by ICML 2026) ---
\section*{Impact Statement}

This paper presents work whose goal is to advance the field of Machine Learning, specifically the detection of artifacts in AI-generated images. Our work could benefit creative image generation, but improved generation quality also increases the risk from deepfakes. We show that our methods also improve AI-generated content detectors, which is one technical mitigation. No personal data is used in model building. Further discussion of ethical considerations, data provenance, and our human labeling study is provided in \cref{sec:appendix-ethics}.
% --- References ---
{
    \small
    \bibliography{main}

@String(ICLR = {Int. Conf. Learn. Represent.})

@String(AAAI = {AAAI})

@String(ICLR  = {ICLR})

@article{ramesh2022hierarchical,
  title={Hierarchical text-conditional image generation with clip latents},
  author={Ramesh, Aditya and Dhariwal, Prafulla and Nichol, Alex and Chu, Casey and Chen, Mark},
  journal={arXiv preprint arXiv:2204.06125},
  volume={1},
  number={2},
  pages={3},
  year={2022}
}

@article{batifol2025flux,
  title={FLUX. 1 Kontext: Flow Matching for In-Context Image Generation and Editing in Latent Space},
  author={Batifol, Stephen and Blattmann, Andreas and Boesel, Frederic and Consul, Saksham and Diagne, Cyril and Dockhorn, Tim and English, Jack and English, Zion and Esser, Patrick and Kulal, Sumith and others},
  journal={arXiv e-prints},
  pages={arXiv--2506},
  year={2025}
}

@article{wu2025qwen,
  title={Qwen-image technical report},
  author={Wu, Chenfei and Li, Jiahao and Zhou, Jingren and Lin, Junyang and Gao, Kaiyuan and Yan, Kun and Yin, Sheng-ming and Bai, Shuai and Xu, Xiao and Chen, Yilei and others},
  journal={arXiv preprint arXiv:2508.02324},
  year={2025}
}

@inproceedings{rombach2022high,
  title={High-resolution image synthesis with latent diffusion models},
  author={Rombach, Robin and Blattmann, Andreas and Lorenz, Dominik and Esser, Patrick and Ommer, Bj{\"o}rn},
  booktitle={Proceedings of the IEEE/CVF conference on computer vision and pattern recognition},
  pages={10684--10695},
  year={2022}
}

@article{cao2024synartifact,
  title={Synartifact: Classifying and alleviating artifacts in synthetic images via vision-language model},
  author={Cao, Bin and Yuan, Jianhao and Liu, Yexin and Li, Jian and Sun, Shuyang and Liu, Jing and Zhao, Bo},
  journal={arXiv preprint arXiv:2402.18068},
  year={2024}
}

@inproceedings{fang2024humanrefiner,
  title={Humanrefiner: Benchmarking abnormal human generation and refining with coarse-to-fine pose-reversible guidance},
  author={Fang, Guian and Yan, Wenbiao and Guo, Yuanfan and Han, Jianhua and Jiang, Zutao and Xu, Hang and Liao, Shengcai and Liang, Xiaodan},
  booktitle={European Conference on Computer Vision},
  pages={201--217},
  year={2024},
  organization={Springer}
}

@article{wang2024detecting,
  title={Detecting Human Artifacts from Text-to-Image Models},
  author={Wang, Kaihong and Zhang, Lingzhi and Zhang, Jianming},
  journal={arXiv preprint arXiv:2411.13842},
  year={2024}
}

@article{wang2025magicmirror,
  title={MagicMirror: A Large-Scale Dataset and Benchmark for Fine-Grained Artifacts Assessment in Text-to-Image Generation},
  author={Wang, Jia and Hu, Jie and Ma, Xiaoqi and Ma, Hanghang and Zeng, Yanbing and Wei, Xiaoming},
  journal={arXiv preprint arXiv:2509.10260},
  year={2025}
}

@inproceedings{wallace2024diffusion,
  title={Diffusion model alignment using direct preference optimization},
  author={Wallace, Bram and Dang, Meihua and Rafailov, Rafael and Zhou, Linqi and Lou, Aaron and Purushwalkam, Senthil and Ermon, Stefano and Xiong, Caiming and Joty, Shafiq and Naik, Nikhil},
  booktitle={Proceedings of the IEEE/CVF Conference on Computer Vision and Pattern Recognition},
  pages={8228--8238},
  year={2024}
}

@article{kynkaanniemi2024applying,
  title={Applying guidance in a limited interval improves sample and distribution quality in diffusion models},
  author={Kynk{\"a}{\"a}nniemi, Tuomas and Aittala, Miika and Karras, Tero and Laine, Samuli and Aila, Timo and Lehtinen, Jaakko},
  journal={Advances in Neural Information Processing Systems},
  volume={37},
  pages={122458--122483},
  year={2024}
}

@article{parmar2025scaling,
  title={Scaling Group Inference for Diverse and High-Quality Generation},
  author={Parmar, Gaurav and Patashnik, Or and Ostashev, Daniil and Wang, Kuan-Chieh and Aberman, Kfir and Narasimhan, Srinivasa and Zhu, Jun-Yan},
  journal={arXiv preprint arXiv:2508.15773},
  year={2025}
}

@article{fan2023dpok,
  title={Dpok: Reinforcement learning for fine-tuning text-to-image diffusion models},
  author={Fan, Ying and Watkins, Olivia and Du, Yuqing and Liu, Hao and Ryu, Moonkyung and Boutilier, Craig and Abbeel, Pieter and Ghavamzadeh, Mohammad and Lee, Kangwook and Lee, Kimin},
  journal={Advances in Neural Information Processing Systems},
  volume={36},
  pages={79858--79885},
  year={2023}
}

@inproceedings{chen2025snapgen,
  title={Snapgen: Taming high-resolution text-to-image models for mobile devices with efficient architectures and training},
  author={Chen, Jierun and Hu, Dongting and Huang, Xijie and Coskun, Huseyin and Sahni, Arpit and Gupta, Aarush and Goyal, Anujraaj and Lahiri, Dishani and Singh, Rajesh and Idelbayev, Yerlan and others},
  booktitle={Proceedings of the Computer Vision and Pattern Recognition Conference},
  pages={7997--8008},
  year={2025}
}

@article{wu2025lightgen,
  title={Lightgen: Efficient image generation through knowledge distillation and direct preference optimization},
  author={Wu, Xianfeng and Bai, Yajing and Zheng, Haoze and Chen, Harold Haodong and Liu, Yexin and Wang, Zihao and Ma, Xuran and Shu, Wen-Jie and Wu, Xianzu and Yang, Harry and others},
  journal={arXiv preprint arXiv:2503.08619},
  year={2025}
}

@inproceedings{nguyen2024swiftbrush,
  title={Swiftbrush: One-step text-to-image diffusion model with variational score distillation},
  author={Nguyen, Thuan Hoang and Tran, Anh},
  booktitle={Proceedings of the IEEE/CVF Conference on Computer Vision and Pattern Recognition},
  pages={7807--7816},
  year={2024}
}

@inproceedings{wang2025generated,
  title={Is this generated person existed in real-world? fine-grained detecting and calibrating abnormal human-body},
  author={Wang, Zeqing and Ma, Qingyang and Wan, Wentao and Li, Haojie and Wang, Keze and Tian, Yonghong},
  booktitle={Proceedings of the Computer Vision and Pattern Recognition Conference},
  pages={21226--21237},
  year={2025}
}

@article{vo2025vision,
  title={Vision Language Models are Biased},
  author={Vo, An and Nguyen, Khai-Nguyen and Taesiri, Mohammad Reza and Dang, Vy Tuong and Nguyen, Anh Totti and Kim, Daeyoung},
  journal={arXiv preprint arXiv:2505.23941},
  year={2025}
}

@inproceedings{goyal2017making,
  title={Making the v in vqa matter: Elevating the role of image understanding in visual question answering},
  author={Goyal, Yash and Khot, Tejas and Summers-Stay, Douglas and Batra, Dhruv and Parikh, Devi},
  booktitle={Proceedings of the IEEE conference on computer vision and pattern recognition},
  pages={6904--6913},
  year={2017}
}

@inproceedings{leng2024mitigating,
  title={Mitigating object hallucinations in large vision-language models through visual contrastive decoding},
  author={Leng, Sicong and Zhang, Hang and Chen, Guanzheng and Li, Xin and Lu, Shijian and Miao, Chunyan and Bing, Lidong},
  booktitle={Proceedings of the IEEE/CVF Conference on Computer Vision and Pattern Recognition},
  pages={13872--13882},
  year={2024}
}

@article{udandarao2024no,
  title={No" zero-shot" without exponential data: Pretraining concept frequency determines multimodal model performance},
  author={Udandarao, Vishaal and Prabhu, Ameya and Ghosh, Adhiraj and Sharma, Yash and Torr, Philip and Bibi, Adel and Albanie, Samuel and Bethge, Matthias},
  journal={Advances in Neural Information Processing Systems},
  volume={37},
  pages={61735--61792},
  year={2024}
}

@article{zhang2024visually,
  title={Why are visually-grounded language models bad at image classification?},
  author={Zhang, Yuhui and Unell, Alyssa and Wang, Xiaohan and Ghosh, Dhruba and Su, Yuchang and Schmidt, Ludwig and Yeung-Levy, Serena},
  journal={Advances in Neural Information Processing Systems},
  volume={37},
  pages={51727--51753},
  year={2024}
}

@inproceedings{yang2023large,
  title={Large language models as optimizers},
  author={Yang, Chengrun and Wang, Xuezhi and Lu, Yifeng and Liu, Hanxiao and Le, Quoc V and Zhou, Denny and Chen, Xinyun},
  booktitle={The Twelfth International Conference on Learning Representations},
  year={2023}
}

@inproceedings{hong2023metagpt,
  title={MetaGPT: Meta programming for a multi-agent collaborative framework},
  author={Hong, Sirui and Zhuge, Mingchen and Chen, Jonathan and Zheng, Xiawu and Cheng, Yuheng and Wang, Jinlin and Zhang, Ceyao and Wang, Zili and Yau, Steven Ka Shing and Lin, Zijuan and others},
  booktitle={The Twelfth International Conference on Learning Representations},
  year={2023}
}

@article{khattab2022demonstrate,
  title={Demonstrate-search-predict: Composing retrieval and language models for knowledge-intensive nlp},
  author={Khattab, Omar and Santhanam, Keshav and Li, Xiang Lisa and Hall, David and Liang, Percy and Potts, Christopher and Zaharia, Matei},
  journal={arXiv preprint arXiv:2212.14024},
  year={2022}
}

@article{alayrac2022flamingo,
  title={Flamingo: a visual language model for few-shot learning},
  author={Alayrac, Jean-Baptiste and Donahue, Jeff and Luc, Pauline and Miech, Antoine and Barr, Iain and Hasson, Yana and Lenc, Karel and Mensch, Arthur and Millican, Katherine and Reynolds, Malcolm and others},
  journal={Advances in neural information processing systems},
  volume={35},
  pages={23716--23736},
  year={2022}
}

@inproceedings{khattab2024dspy,
  title={DSPy: Compiling Declarative Language Model Calls into Self-Improving Pipelines},
  author={Khattab, Omar and Singhvi, Arnav and Maheshwari, Paridhi and Zhang, Zhiyuan and Santhanam, Keshav and Vardhamanan, Sri and Haq, Saiful and Sharma, Ashutosh and Joshi, Thomas T. and Moazam, Hanna and Miller, Heather and Zaharia, Matei and Potts, Christopher},
  booktitle={The Twelfth International Conference on Learning Representations (ICLR)},
  year={2024},
}

@article{zhou2023lima,
  title={Lima: Less is more for alignment},
  author={Zhou, Chunting and Liu, Pengfei and Xu, Puxin and Iyer, Srinivasan and Sun, Jiao and Mao, Yuning and Ma, Xuezhe and Efrat, Avia and Yu, Ping and Yu, Lili and others},
  journal={Advances in Neural Information Processing Systems},
  volume={36},
  pages={55006--55021},
  year={2023}
}

@inproceedings{doveh2024towards,
  title={Towards Multimodal In-context Learning for Vision and Language Models},
  author={Doveh, Sivan and Perek, Shaked and Mirza, M Jehanzeb and Lin, Wei and Alfassy, Amit and Arbelle, Assaf and Ullman, Shimon and Karlinsky, Leonid},
  booktitle={European Conference on Computer Vision},
  pages={250--267},
  year={2024},
  organization={Springer}
}

@misc{compound-ai-blog,
  title        = {The Shift from Models to Compound AI Systems},
  author       = {Matei Zaharia and Omar Khattab and Lingjiao Chen and Jared Quincy Davis and Heather Miller and Chris Potts and James Zou and Michael Carbin and Jonathan Frankle and Naveen Rao and Ali Ghodsi},
  howpublished = {\url{https://bair.berkeley.edu/blog/2024/02/18/compound-ai-systems/}},
  year         = {2024}
}

@misc{blackforestlabs2024flux,
  author       = {Black Forest Labs},
  title        = {Flux: Official inference repository for FLUX.1 models},
  year         = {2024},
  howpublished = {\url{https://github.com/black-forest-labs/flux}},
  note         = {GitHub repository}
}

@misc{gemini25pro,
  title        = {Gemini 2.5 Pro: Pushing the Frontier with Advanced Reasoning},
  author       = {{Google DeepMind / Google Cloud}},
  year         = {2025},
  howpublished = {\url{https://storage.googleapis.com/deepmind-media/gemini/gemini_v2_5_report.pdf}},
  note         = {Technical report; part of the Gemini 2.X model family.},
}

@misc{gpt4o,
  title        = {GPT-4o System Card},
  author       = {{OpenAI}},
  year         = {2024},
  howpublished = {\url{https://openai.com/index/gpt-4o-system-card/}},
  note         = {Multimodal “omni” model accepting text, audio, image, video.},
}

@misc{gpt5,
  title        = {GPT-5 System Card},
  author       = {{OpenAI}},
  year         = {2025},
  howpublished = {\url{https://cdn.openai.com/gpt-5-system-card.pdf}},
  note         = {Unified system with routing across model variants.},
}

@misc{qwen2_5_vl,
  title        = {Qwen 2.5-VL: A Vision-Language Model Series},
  author       = {{Alibaba Cloud / Qwen Team}},
  year         = {2025},
  howpublished = {\url{https://arxiv.org/abs/2502.13923}},
  note         = {Multimodal model series (image/video + text) — e.g. “Qwen2.5-VL”.}  
}

@article{liu2024chain,
  title={Chain-of-spot: Interactive reasoning improves large vision-language models},
  author={Liu, Zuyan and Dong, Yuhao and Rao, Yongming and Zhou, Jie and Lu, Jiwen},
  journal={arXiv preprint arXiv:2403.12966},
  year={2024}
}

@inproceedings{wu2024v,
  title={V?: Guided visual search as a core mechanism in multimodal llms},
  author={Wu, Penghao and Xie, Saining},
  booktitle={Proceedings of the IEEE/CVF Conference on Computer Vision and Pattern Recognition},
  pages={13084--13094},
  year={2024}
}

@article{shao2024visual,
  title={Visual cot: Advancing multi-modal language models with a comprehensive dataset and benchmark for chain-of-thought reasoning},
  author={Shao, Hao and Qian, Shengju and Xiao, Han and Song, Guanglu and Zong, Zhuofan and Wang, Letian and Liu, Yu and Li, Hongsheng},
  journal={Advances in Neural Information Processing Systems},
  volume={37},
  pages={8612--8642},
  year={2024}
}

@inproceedings{liu2024grounding,
  title={Grounding dino: Marrying dino with grounded pre-training for open-set object detection},
  author={Liu, Shilong and Zeng, Zhaoyang and Ren, Tianhe and Li, Feng and Zhang, Hao and Yang, Jie and Jiang, Qing and Li, Chunyuan and Yang, Jianwei and Su, Hang and others},
  booktitle={European conference on computer vision},
  pages={38--55},
  year={2024},
  organization={Springer}
}

@inproceedings{sun2024generative,
  title={Generative multimodal models are in-context learners},
  author={Sun, Quan and Cui, Yufeng and Zhang, Xiaosong and Zhang, Fan and Yu, Qiying and Wang, Yueze and Rao, Yongming and Liu, Jingjing and Huang, Tiejun and Wang, Xinlong},
  booktitle={Proceedings of the IEEE/CVF Conference on Computer Vision and Pattern Recognition},
  pages={14398--14409},
  year={2024}
}

@inproceedings{yang2023re,
  title={Re-vilm: Retrieval-augmented visual language model for zero and few-shot image captioning},
  author={Yang, Zhuolin and Ping, Wei and Liu, Zihan and Korthikanti, Vijay and Nie, Weili and Huang, De-An and Fan, Linxi and Yu, Zhiding and Lan, Shiyi and Li, Bo and others},
  booktitle={Findings of the Association for Computational Linguistics: EMNLP 2023},
  pages={11844--11857},
  year={2023}
}

@inproceedings{radford2021learning,
  title={Learning transferable visual models from natural language supervision},
  author={Radford, Alec and Kim, Jong Wook and Hallacy, Chris and Ramesh, Aditya and Goh, Gabriel and Agarwal, Sandhini and Sastry, Girish and Askell, Amanda and Mishkin, Pamela and Clark, Jack and others},
  booktitle={International conference on machine learning},
  pages={8748--8763},
  year={2021},
  organization={PmLR}
}

@inproceedings{zhou2022large,
  title={Large language models are human-level prompt engineers},
  author={Zhou, Yongchao and Muresanu, Andrei Ioan and Han, Ziwen and Paster, Keiran and Pitis, Silviu and Chan, Harris and Ba, Jimmy},
  booktitle={The eleventh international conference on learning representations},
  year={2022}
}

@article{schulhoff2024prompt,
  title={The prompt report: a systematic survey of prompt engineering techniques},
  author={Schulhoff, Sander and Ilie, Michael and Balepur, Nishant and Kahadze, Konstantine and Liu, Amanda and Si, Chenglei and Li, Yinheng and Gupta, Aayush and Han, HyoJung and Schulhoff, Sevien and others},
  journal={arXiv preprint arXiv:2406.06608},
  year={2024}
}

@inproceedings{gandikota2024concept,
  title={Concept sliders: Lora adaptors for precise control in diffusion models},
  author={Gandikota, Rohit and Materzy{\'n}ska, Joanna and Zhou, Tingrui and Torralba, Antonio and Bau, David},
  booktitle={European Conference on Computer Vision},
  pages={172--188},
  year={2024},
  organization={Springer}
}

@inproceedings{lu2024handrefiner,
  title={Handrefiner: Refining malformed hands in generated images by diffusion-based conditional inpainting},
  author={Lu, Wenquan and Xu, Yufei and Zhang, Jing and Wang, Chaoyue and Tao, Dacheng},
  booktitle={Proceedings of the 32nd ACM International Conference on Multimedia},
  pages={7085--7093},
  year={2024}
}

@article{wang2025diffdoctor,
  title={DiffDoctor: Diagnosing Image Diffusion Models Before Treating},
  author={Wang, Yiyang and Chen, Xi and Xu, Xiaogang and Ji, Sihui and Liu, Yu and Shen, Yujun and Zhao, Hengshuang},
  journal={arXiv preprint arXiv:2501.12382},
  year={2025}
}

@inproceedings{wang2025rhands,
  title={Rhands: Refining malformed hands for generated images with decoupled structure and style guidance},
  author={Wang, Chengrui and Liu, Pengfei and Zhou, Min and Zeng, Ming and Li, Xubin and Ge, Tiezheng and Zheng, Bo},
  booktitle={Proceedings of the AAAI Conference on Artificial Intelligence},
  volume={39},
  number={7},
  pages={7573--7581},
  year={2025}
}

@article{zhou2025aigi,
  title={AIGI-Holmes: Towards Explainable and Generalizable AI-Generated Image Detection via Multimodal Large Language Models},
  author={Zhou, Ziyin and Luo, Yunpeng and Wu, Yuanchen and Sun, Ke and Ji, Jiayi and Yan, Ke and Ding, Shouhong and Sun, Xiaoshuai and Wu, Yunsheng and Ji, Rongrong},
  journal={arXiv preprint arXiv:2507.02664},
  year={2025}
}

@inproceedings{wang2020cnn,
  title={CNN-generated images are surprisingly easy to spot... for now},
  author={Wang, Sheng-Yu and Wang, Oliver and Zhang, Richard and Owens, Andrew and Efros, Alexei A},
  booktitle={Proceedings of the IEEE/CVF conference on computer vision and pattern recognition},
  pages={8695--8704},
  year={2020}
}

@inproceedings{tan2024rethinking,
  title={Rethinking the up-sampling operations in cnn-based generative network for generalizable deepfake detection},
  author={Tan, Chuangchuang and Zhao, Yao and Wei, Shikui and Gu, Guanghua and Liu, Ping and Wei, Yunchao},
  booktitle={Proceedings of the IEEE/CVF Conference on Computer Vision and Pattern Recognition},
  pages={28130--28139},
  year={2024}
}

@inproceedings{luo2024lare,
  title={LaRE\^{} 2: Latent reconstruction error based method for diffusion-generated image detection},
  author={Luo, Yunpeng and Du, Junlong and Yan, Ke and Ding, Shouhong},
  booktitle={Proceedings of the IEEE/CVF Conference on Computer Vision and Pattern Recognition},
  pages={17006--17015},
  year={2024}
}

@article{li2025fakebench,
  title={Fakebench: Probing explainable fake image detection via large multimodal models},
  author={Li, Yixuan and Liu, Xuelin and Wang, Xiaoyang and Lee, Bu Sung and Wang, Shiqi and Rocha, Anderson and Lin, Weisi},
  journal={IEEE Transactions on Information Forensics and Security},
  year={2025},
  publisher={IEEE}
}

@article{zhu2023genimage,
  title={Genimage: A million-scale benchmark for detecting ai-generated image},
  author={Zhu, Mingjian and Chen, Hanting and Yan, Qiangyu and Huang, Xudong and Lin, Guanyu and Li, Wei and Tu, Zhijun and Hu, Hailin and Hu, Jie and Wang, Yunhe},
  journal={Advances in Neural Information Processing Systems},
  volume={36},
  pages={77771--77782},
  year={2023}
}

@article{xu2023imagereward,
  title={Imagereward: Learning and evaluating human preferences for text-to-image generation},
  author={Xu, Jiazheng and Liu, Xiao and Wu, Yuchen and Tong, Yuxuan and Li, Qinkai and Ding, Ming and Tang, Jie and Dong, Yuxiao},
  journal={Advances in Neural Information Processing Systems},
  volume={36},
  pages={15903--15935},
  year={2023}
}

@article{wu2023human,
  title={Human preference score v2: A solid benchmark for evaluating human preferences of text-to-image synthesis},
  author={Wu, Xiaoshi and Hao, Yiming and Sun, Keqiang and Chen, Yixiong and Zhu, Feng and Zhao, Rui and Li, Hongsheng},
  journal={arXiv preprint arXiv:2306.09341},
  year={2023}
}

@article{kang2025geneva,
  title={GeneVA: A Dataset of Human Annotations for Generative Text to Video Artifacts},
  author={Kang, Jenna and Silva, Maria and Sangkloy, Patsorn and Chen, Kenneth and Williams, Niall and Sun, Qi},
  journal={arXiv preprint arXiv:2509.08818},
  year={2025}
}

@article{wu2025visualquality,
  title={VisualQuality-R1: Reasoning-Induced Image Quality Assessment via Reinforcement Learning to Rank},
  author={Wu, Tianhe and Zou, Jian and Liang, Jie and Zhang, Lei and Ma, Kede},
  journal={arXiv preprint arXiv:2505.14460},
  year={2025}
}

@article{li2025q,
  title={Q-insight: Understanding image quality via visual reinforcement learning},
  author={Li, Weiqi and Zhang, Xuanyu and Zhao, Shijie and Zhang, Yabin and Li, Junlin and Zhang, Li and Zhang, Jian},
  journal={arXiv preprint arXiv:2503.22679},
  year={2025}
}

@article{wu2023q,
  title={Q-align: Teaching lmms for visual scoring via discrete text-defined levels},
  author={Wu, Haoning and Zhang, Zicheng and Zhang, Weixia and Chen, Chaofeng and Liao, Liang and Li, Chunyi and Gao, Yixuan and Wang, Annan and Zhang, Erli and Sun, Wenxiu and others},
  journal={arXiv preprint arXiv:2312.17090},
  year={2023}
}

@inproceedings{you2024depicting,
  title={Depicting beyond scores: Advancing image quality assessment through multi-modal language models},
  author={You, Zhiyuan and Li, Zheyuan and Gu, Jinjin and Yin, Zhenfei and Xue, Tianfan and Dong, Chao},
  booktitle={European Conference on Computer Vision},
  pages={259--276},
  year={2024},
  organization={Springer}
}

@misc{anthropic2024effectiveagents,
  author       = {Anthropic},
  title        = {Building Effective Agents},
  year         = {2024},
  month        = {December},
  howpublished = {\url{https://www.anthropic.com/engineering/building-effective-agents}},
  note         = {Accessed: 2025-11-10}
}

@inproceedings{wu2024autogen,
  title={Autogen: Enabling next-gen LLM applications via multi-agent conversations},
  author={Wu, Qingyun and Bansal, Gagan and Zhang, Jieyu and Wu, Yiran and Li, Beibin and Zhu, Erkang and Jiang, Li and Zhang, Xiaoyun and Zhang, Shaokun and Liu, Jiale and others},
  booktitle={First Conference on Language Modeling},
  year={2024}
}

@inproceedings{yao2022react,
  title={React: Synergizing reasoning and acting in language models},
  author={Yao, Shunyu and Zhao, Jeffrey and Yu, Dian and Du, Nan and Shafran, Izhak and Narasimhan, Karthik R and Cao, Yuan},
  booktitle={The eleventh international conference on learning representations},
  year={2022}
}

@inproceedings{du2023improving,
  title={Improving factuality and reasoning in language models through multiagent debate},
  author={Du, Yilun and Li, Shuang and Torralba, Antonio and Tenenbaum, Joshua B and Mordatch, Igor},
  booktitle={Forty-first International Conference on Machine Learning},
  year={2023}
}

@article{wu2025optimas,
  title={Optimas: Optimizing compound ai systems with globally aligned local rewards},
  author={Wu, Shirley and Sarthi, Parth and Zhao, Shiyu and Lee, Aaron and Shandilya, Herumb and Grobelnik, Adrian Mladenic and Choudhary, Nurendra and Huang, Eddie and Subbian, Karthik and Zhang, Linjun and others},
  journal={arXiv preprint arXiv:2507.03041},
  year={2025}
}

@article{huang2024multimodal,
  title={Multimodal task vectors enable many-shot multimodal in-context learning},
  author={Huang, Brandon and Mitra, Chancharik and Arbelle, Assaf and Karlinsky, Leonid and Darrell, Trevor and Herzig, Roei},
  journal={Advances in Neural Information Processing Systems},
  volume={37},
  pages={22124--22153},
  year={2024}
}

@article{brown2020language,
  title={Language models are few-shot learners},
  author={Brown, Tom and Mann, Benjamin and Ryder, Nick and Subbiah, Melanie and Kaplan, Jared D and Dhariwal, Prafulla and Neelakantan, Arvind and Shyam, Pranav and Sastry, Girish and Askell, Amanda and others},
  journal={Advances in neural information processing systems},
  volume={33},
  pages={1877--1901},
  year={2020}
}

@article{zong2024vl,
  title={VL-ICL bench: The devil in the details of multimodal in-context learning},
  author={Zong, Yongshuo and Bohdal, Ondrej and Hospedales, Timothy},
  journal={arXiv preprint arXiv:2403.13164},
  year={2024}
}

@article{qin2024factors,
  title={What factors affect multi-modal in-context learning? an in-depth exploration},
  author={Qin, Libo and Chen, Qiguang and Fei, Hao and Chen, Zhi and Li, Min and Che, Wanxiang},
  journal={Advances in Neural Information Processing Systems},
  volume={37},
  pages={123207--123236},
  year={2024}
}

@article{jiang2024many,
  title={Many-shot in-context learning in multimodal foundation models},
  author={Jiang, Yixing and Irvin, Jeremy and Wang, Ji Hun and Chaudhry, Muhammad Ahmed and Chen, Jonathan H and Ng, Andrew Y},
  journal={arXiv preprint arXiv:2405.09798},
  year={2024}
}

@article{agrawal2025gepa,
  title={Gepa: Reflective prompt evolution can outperform reinforcement learning},
  author={Agrawal, Lakshya A and Tan, Shangyin and Soylu, Dilara and Ziems, Noah and Khare, Rishi and Opsahl-Ong, Krista and Singhvi, Arnav and Shandilya, Herumb and Ryan, Michael J and Jiang, Meng and others},
  journal={arXiv preprint arXiv:2507.19457},
  year={2025}
}

@inproceedings{bergmann2019mvtec,
  title={MVTec AD--A comprehensive real-world dataset for unsupervised anomaly detection},
  author={Bergmann, Paul and Fauser, Michael and Sattlegger, David and Steger, Carsten},
  booktitle={Proceedings of the IEEE/CVF conference on computer vision and pattern recognition},
  pages={9592--9600},
  year={2019}
}

@article{hendrycks2018deep,
  title={Deep anomaly detection with outlier exposure},
  author={Hendrycks, Dan and Mazeika, Mantas and Dietterich, Thomas},
  journal={arXiv preprint arXiv:1812.04606},
  year={2018}
}

@article{xiang2025self,
  title={Self-supervised prompt optimization},
  author={Xiang, Jinyu and Zhang, Jiayi and Yu, Zhaoyang and Liang, Xinbing and Teng, Fengwei and Tu, Jinhao and Ren, Fashen and Tang, Xiangru and Hong, Sirui and Wu, Chenglin and others},
  journal={arXiv preprint arXiv:2502.06855},
  year={2025}
}

@article{achiam2023gpt,
  title={Gpt-4 technical report},
  author={Achiam, Josh and Adler, Steven and Agarwal, Sandhini and Ahmad, Lama and Akkaya, Ilge and Aleman, Florencia Leoni and Almeida, Diogo and Altenschmidt, Janko and Altman, Sam and Anadkat, Shyamal and others},
  journal={arXiv preprint arXiv:2303.08774},
  year={2023}
}

@inproceedings{recht2019imagenet,
  title={Do imagenet classifiers generalize to imagenet?},
  author={Recht, Benjamin and Roelofs, Rebecca and Schmidt, Ludwig and Shankar, Vaishaal},
  booktitle={International conference on machine learning},
  pages={5389--5400},
  year={2019},
  organization={PMLR}
}

@article{michaud2024elicitation,
  title        = {Elicitation: The Simplest Way to Understand Post-Training},
  author       = {Nathan Lambert},
  year         = {2024},
  journal      = {Interconnects},
  url          = {https://www.interconnects.ai/p/elicitation-theory-of-post-training},
  note         = {Accessed: 2025-11-11}
}

@article{cohen1960coefficient,
  title={A coefficient of agreement for nominal scales},
  author={Cohen, Jacob},
  journal={Educational and psychological measurement},
  volume={20},
  number={1},
  pages={37--46},
  year={1960},
  publisher={Sage Publications Sage CA: Thousand Oaks, CA}
}

@article{landis1977measurement,
  title={The measurement of observer agreement for categorical data},
  author={Landis, J Richard and Koch, Gary G},
  journal={biometrics},
  pages={159--174},
  year={1977},
  publisher={JSTOR}
}

@article{chen2022program,
  title={Program of thoughts prompting: Disentangling computation from reasoning for numerical reasoning tasks},
  author={Chen, Wenhu and Ma, Xueguang and Wang, Xinyi and Cohen, William W},
  journal={arXiv preprint arXiv:2211.12588},
  year={2022}
}

@article{mirza2024glov,
  title={Glov: Guided large language models as implicit optimizers for vision language models},
  author={Mirza, M Jehanzeb and Zhao, Mengjie and Mao, Zhuoyuan and Doveh, Sivan and Lin, Wei and Gavrikov, Paul and Dorkenwald, Michael and Yang, Shiqi and Jha, Saurav and Wakaki, Hiromi and others},
  journal={arXiv preprint arXiv:2410.06154},
  year={2024}
}

@article{du2024ipo,
  title={Ipo: Interpretable prompt optimization for vision-language models},
  author={Du, Yingjun and Sun, Wenfang and Snoek, Cees},
  journal={Advances in Neural Information Processing Systems},
  volume={37},
  pages={126725--126766},
  year={2024}
}

@article{choi2025multimodal,
  title={Multimodal Prompt Optimization: Why Not Leverage Multiple Modalities for MLLMs},
  author={Choi, Yumin and Kim, Dongki and Baek, Jinheon and Hwang, Sung Ju},
  journal={arXiv preprint arXiv:2510.09201},
  year={2025}
}

@inproceedings{wu2025hierarchical,
  title={Hierarchical Variational Test-Time Prompt Generation for Zero-Shot Generalization},
  author={Wu, Zhaoyang and Liu, Fang and Jiao, Licheng and Li, Shuo and Li, Lingling and Liu, Xu and Chen, Puhua and Ma, Wenping},
  booktitle={Proceedings of the IEEE/CVF International Conference on Computer Vision},
  pages={2325--2335},
  year={2025}
}

@article{shinn2023reflexion,
  title={Reflexion: Language agents with verbal reinforcement learning},
  author={Shinn, Noah and Cassano, Federico and Gopinath, Ashwin and Narasimhan, Karthik and Yao, Shunyu},
  journal={Advances in Neural Information Processing Systems},
  volume={36},
  pages={8634--8652},
  year={2023}
}

@article{fernando2023promptbreeder,
  title={Promptbreeder: Self-referential self-improvement via prompt evolution},
  author={Fernando, Chrisantha and Banarse, Dylan and Michalewski, Henryk and Osindero, Simon and Rockt{\"a}schel, Tim},
  journal={arXiv preprint arXiv:2309.16797},
  year={2023}
}
    \bibliographystyle{icml2026}
}
% --- Appendix ---
\newpage
\appendix
\onecolumn
% Appendix (lettered numbering handled by \appendix in main.tex)

\section{Acknowledgments}
Thanks to Yuhui Zhang and Sivan Doveh for discussions.

\section{Ethics}
\label{sec:appendix-ethics}
This section expands on the Impact Statement in the main paper with additional details on data provenance, societal risks, and our human labeling study.

Regarding data, we only use existing benchmarks proposed by prior works.  We have checked the source websites for each and confirm that none have been withdrawn. 

Regarding societal risks, we note that our work is a continuation of existing work building detection models, which aims to improve the quality and realism of creative image generation models. All such work adds to the risk from deepfakes that humans believe to be authentic, especially for misinformation or fraud. One technical strategy for mitigation is to further develop methods for AI-generated content-detection \cite{zhou2025aigi}, and actually we show that our work improves such models.

No personal data is used in model building. The only human subject involvement was a labeling study ( \cref{sec:results_main_results}) in which a team of ten humans were hired to label 600 samples with `artifact' or `no\_artifact' for human images. This study involved no personal information collection, only professional judgment on publicly available synthetic images. IRB approval was not required for this type of compensated image annotation task that poses no risk to participants.

% \section{Code release}
% We will Public versions will link to a future GitHub repository.

\section{Benchmark details and release}
\label{sec:appendix-benchmark_details}
We test with five existing benchmarks for human artifact detection: SynArtifact \cite{cao2024synartifact}, AbHuman \cite{fang2024humanrefiner}, Human Artifact Detection (HAD) dataset \cite{wang2024detecting}, AIGC-HA-1k \cite{wang2025generated}, and MagicBench \cite{wang2025magicmirror}. No prior work tests on more than one, and so this is the first time that generalization across datasets is evaluated. This section will discuss further details and dataset access instructions.

% \Cref{sec:appendix-benchmark-attributes} describes the benchmarks in more detail, and \Cref{sec:appendix-benchmark-access} describes dataset access.
\subsection{Sample images and error types}
\begin{figure}[ht]
    \centering
    \includegraphics[page=8, width=\columnwidth,  height=\textheight,
    keepaspectratio,
    clip,
    trim=0 0 1140 0
    ]{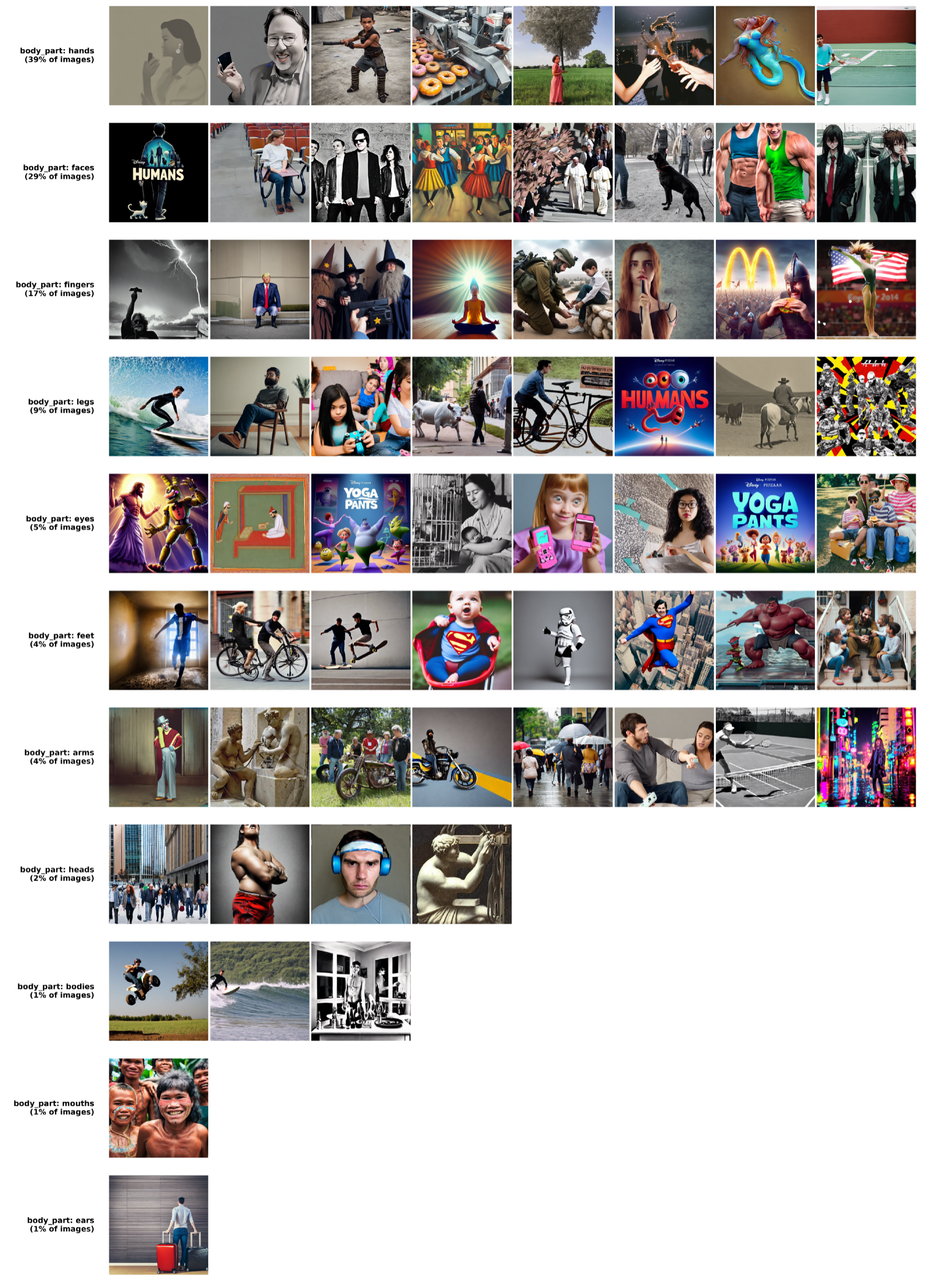}
    \captionof{figure}{
    Sample images per error type in SynArtifact \cite{cao2024synartifact}, ordered by prevalence.
}
    \label{fig:supp-samples-synartifact}
\end{figure}

\begin{figure}[ht]
    \centering
    \includegraphics[page=9, width=\columnwidth,  height=\textheight,
    keepaspectratio,
    clip,
    trim=0 0 1140 0
    ]{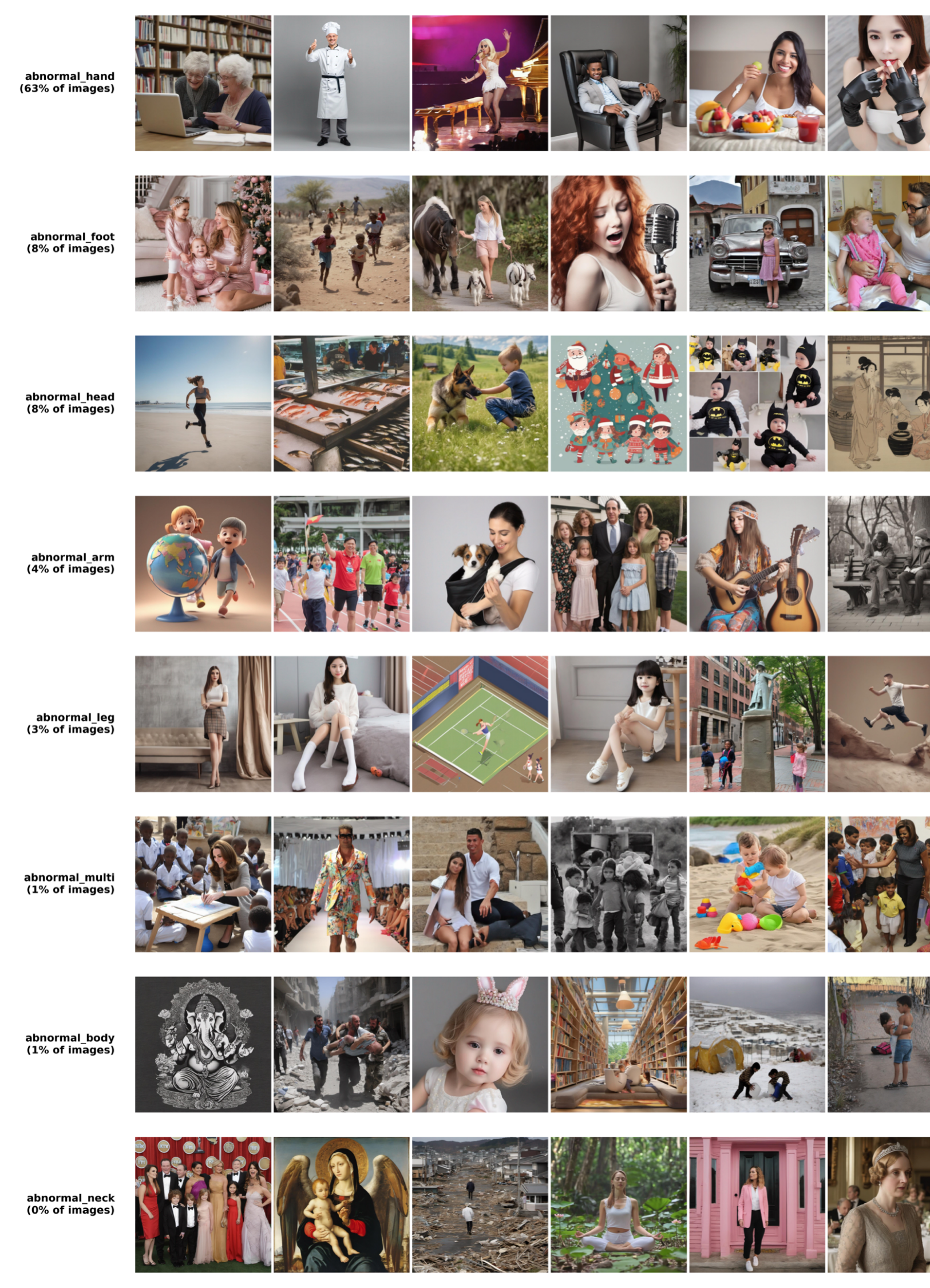}
    \captionof{figure}{
    Sample images per error type in AbHuman \cite{fang2024humanrefiner}, ordered by prevalence.
}
    \label{fig:supp-samples-abhuman}
\end{figure}

\begin{figure}[ht]
    \centering
    \includegraphics[page=10, width=\columnwidth,  height=\textheight,
    keepaspectratio,
    clip,
    trim=0 0 900 0
    ]{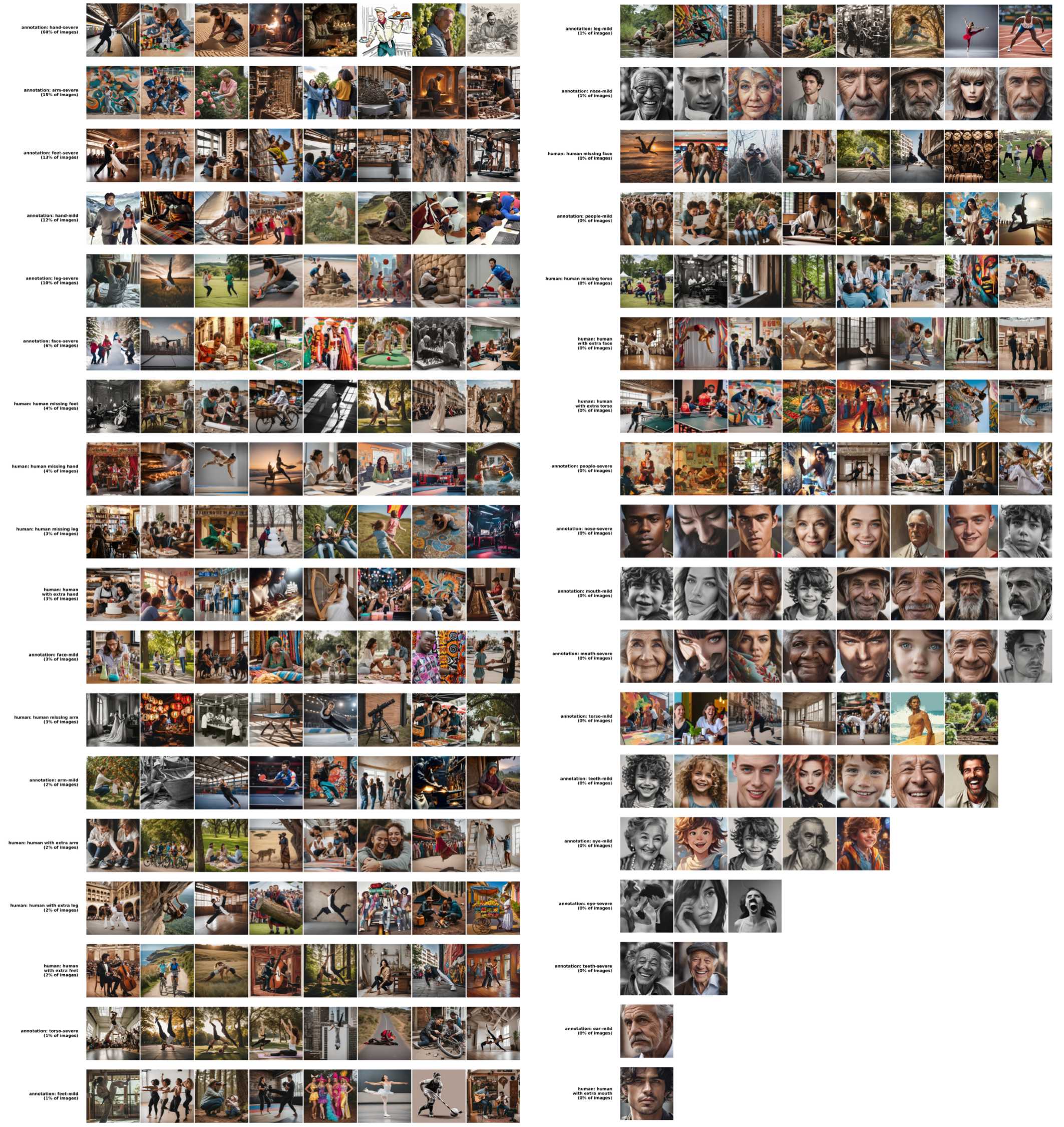}
    \captionof{figure}{
    Sample images per error type in Human Artifact Dataset (HAD) \cite{wang2024detecting}, ordered by prevalence.
}
    \label{fig:supp-samples-had}
\end{figure}

\begin{figure}[ht]
    \centering
    \includegraphics[page=11, width=\columnwidth,  height=\textheight,
    keepaspectratio,
    clip,
    trim=0 0 1240 0
    ]{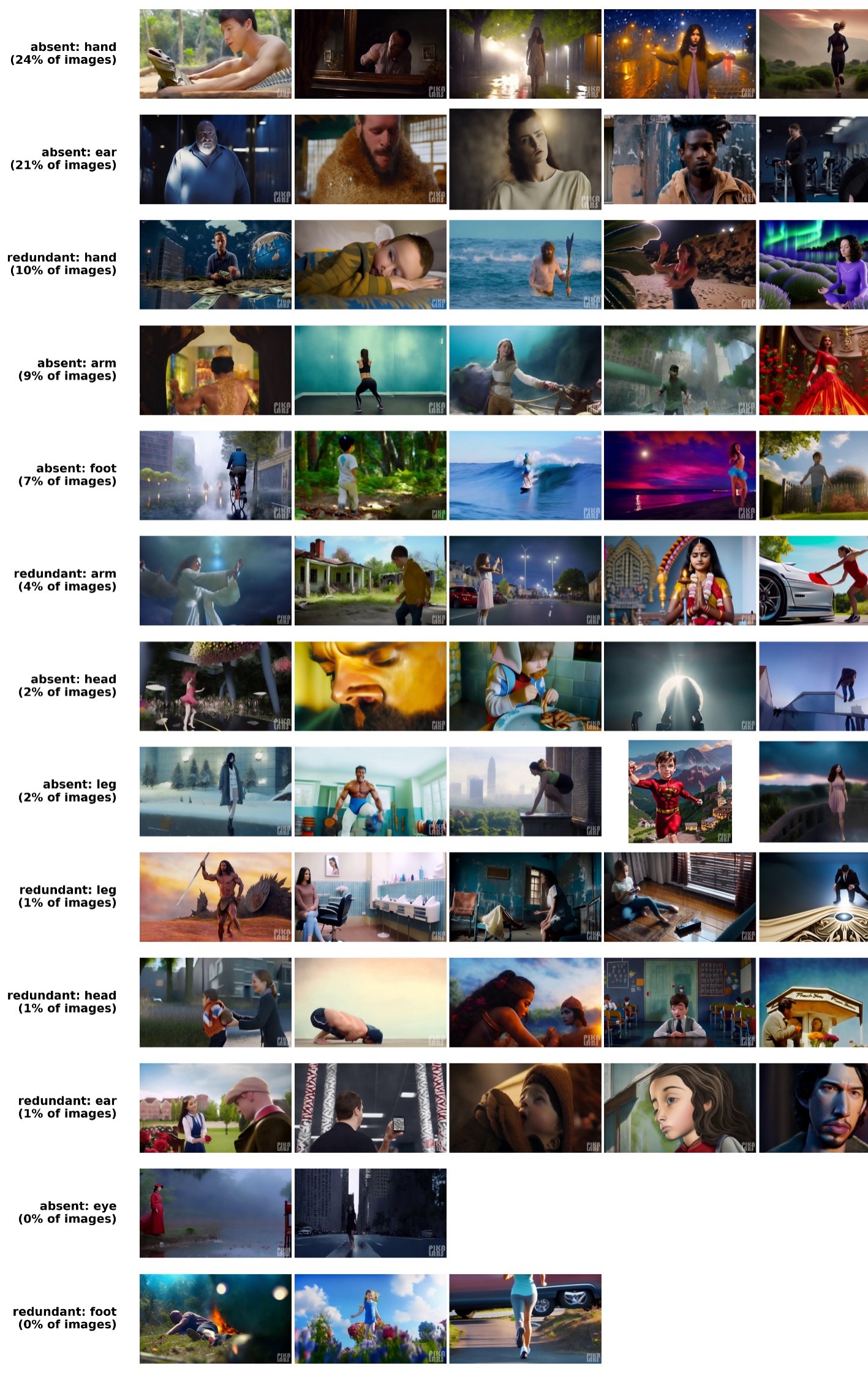}
    \captionof{figure}{
    Sample images per error type in AIGC-Human-Aware 1k \cite{wang2025generated}, ordered by prevalence.
}
    \label{fig:supp-samples-aigc-ha}
\end{figure}

\begin{figure}[ht]
    \centering
    \includegraphics[page=12, width=\columnwidth,  height=\textheight,
    keepaspectratio,
    clip,
    trim=0 0 700 0
    ]{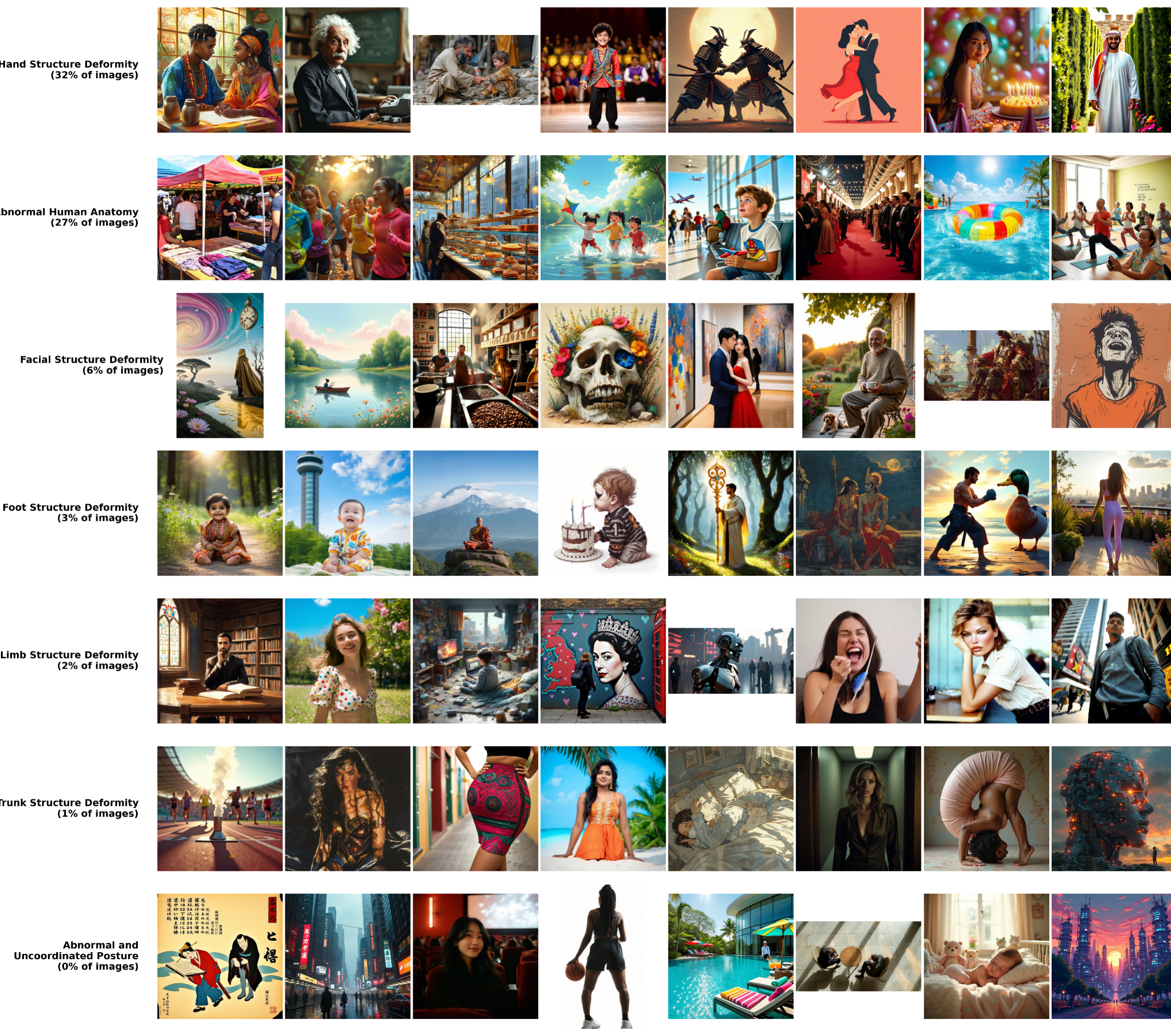}
    \captionof{figure}{
    Sample images per error type in MagicBench \cite{wang2025magicmirror}, ordered by prevalence.
}
    \label{fig:supp-samples-magicbench}
\end{figure}

For each benchmark, we show a grid of images where each row is an artifact class. They are at the end of this document in \cref{fig:supp-samples-synartifact}, 
\cref{fig:supp-samples-abhuman},
\cref{fig:supp-samples-had},
\cref{fig:supp-samples-aigc-ha}, and
\cref{fig:supp-samples-magicbench}.

\subsection{Benchmark attribute details}
\label{sec:appendix-benchmark-attributes}
The summary table in \cref{table:benchmark-attributes} covers the most important attributes, but we provide more details here. Note that in \cref{table:benchmark-attributes}, we have numbers for `number of sublabel types' and `number of samples'; these are \textit{after} doing filtering of non-human images, which we do for SynArtifact, AbHuman, and MagicBench (as we discuss below).

\noindent \textbf{SynArtifact} \cite{cao2024synartifact} 

This is the earliest dataset for image generation artifacts.  It includes both human and non-human artifacts, with 13 classes in its taxonomy: `illegible letters', `awkward facial expression',`distorted  or deformated component' (sic.), `duplicated component', `omitted component', `chromatic irregularity', `abnormal  non spatial relationship', `abnormal spatial relationship' ,`abnormal texture',`luminance discrepancy',`impractical luminosity',`localized blur'.  Many error types include both human- and non-human- types, for example `duplicated component' could be an extra human arm, or an extra chair leg. Since the dataset includes text descriptions for each image label, we identify which labels are human errors by string-matching the caption:
{
\vspace{-0.7\baselineskip}
\footnotesize \begin{verbatim}
['hand', 'finger', 'face', 'eye', 'facial', 'arm', 'leg', 'body', 'head', 
'mouth', 'nose', 'ear', 'foot', 'feet', 'neck', 'torso', 'shoulder']
\end{verbatim}
\vspace{-0.7\baselineskip}
}
From those results, we create a new error taxonomy error classes for `hand', `face', etc. These error classes can be leveraged by \SystemName; however many classes are rare (feet, heads, arms, bodies, ears, noses, mouths), so we merge those classes. This approach discards the error-type (extra body part vs missing body part), so other work could choose to recover it.

Note that creating error classes this way can actually change the overall image label --- for example, an image containing `illegible letters' but a well-generated human does not have any human artifact, and so `artifact=0'.

As discussed in the main paper, we filter images without humans by using a VLM query. We use GPT-4o \cite{achiam2023gpt} and we use the \cref{pr:filter-for-humans} (below). Note that this is the base prompt which is the `Signature' for a simple DSPy program; this means the final prompt has some extra text scaffolding that ensures the output format is properly adhered to.

\FloatBarrier
\begin{promptenv}
\caption{Detecting whether a human body part appears in an image. Images without humans are filtered in our datasets. (DSPy formats this with the image to be analyzed.}
\label{pr:filter-for-humans}
\begin{prompt}
Look at this image and determine if it contains any of these body parts: a human, any human body part, a human hand, a human face, a human foot. Return 1 if at least one of these parts is clearly visible and recognizable, otherwise return 0. You only need ONE of the specified parts to be present to return 1.
\end{prompt}
\end{promptenv}
\FloatBarrier

For generating the original images, they have a complex mix of prompt dataset and image generative model. StableDiffusion v2.1 is used with the prompt sources of ImageNet, MSCOCO, DrawBench, and Midjourney Users. StableDiffusion is used with prompts from ImageReward. DALLE-3 is used with prompts from DALLE-3 users. DrawBench is used again for StableDiffusion versions 2.0, 2.1, 1.0, 1.4, and 1.5.

\noindent \textbf{AbHuman} \cite{fang2024humanrefiner} 

AbHuman has an error taxonomy with seven types: `abnormal head', `abnormal neck', `abnormal body (torso)', `abnormal arm', `abnormal hand', `abnormal leg', `abnormal foot'. They also have classes for `normal head', `normal neck' and so on, which is possible to label because annotations include boxes, but we do not use this information.

There is also a class `not human'. We remove it because this is an easy task for regular VLMs, and because it does not fit properly with our formulation of the task. As such, we filter all images not containing humans, using the same prompt as in `SynArtifact' (\cref{pr:filter-for-humans}).

\noindent \textbf{Human Artifact Dataset (HAD)} \cite{wang2024detecting}
HAD contains two types of error; in their labels, they are called `human' and `annotation', but we call them `coarse' and `fine' errors. 

For coarse errors, they identify eight body parts: arm, feet, hand, leg, face, mouth, nose, and torso. There is an error class for both `missing' and `extra', so there is `missing arm', `extra arm', `missing feet', `extra feet', and so on for sixteen label types. These obviously refer to parts that have extras or are missing. 

For `fine' errors, they consider 12 parts: arm, face, feet, hand, leg, torso, ear, eye, mouth, nose, teeth, and people (the `people' class is when a deformity cannot be easily localized to one part, though it is not common). For each there is a class for `severe' and `mild' artifact, for example `arm-severe', `arm-mild', `face-severe', `face-mild', and so on. These correspond to `deformities' though the body part does exist. 

When defining the sub-labels for our methods, we merge rare classes. We also combine the `severe' and `mild' classes, for example `face-mild' and `face-severe' are merged to `abnormal-face'.

While it is a good dataset overall, an issue we find in HAD is that the images are saved with resolution 512, which is small when trying to evaluate small visual regions. 

\noindent \textbf{AIGC Human-Aware 1K (AIGC-HA)} \cite{wang2025generated}

Here, AIGC stands for `AI-generated content detection'. For error classes, they consider these parts: hand, arm, leg, foot, head, ear, eye. For each there is a class for whether it is `extra' or `redundant', for example there is `extra hand', `redundant hand', `extra arm', `redundant arm', and so on. Different other benchmarks, they only consider extra or missing parts, and not \textit{deformed} parts. However, by inspecting data samples, we find this gap is not as great as it seems. For example, there are images marked as `missing hand' which, under the labeling rules of other benchmarks, would be labeled as `deformed hand'. The taxonomy is therefore similar to the others.

Similar to the other benchmarks, AIGC-HA generates images with text-to-image models and annotations are performed by humans.  However, different to the other benchmarks, the attached training set is not from the same distribution. Instead, it is synthetically generated -- the advantage is that they can more scalably generate data, but the disadvantage is the distribution shift. The synthetic data is only for the `missing part' classes. They take COCO images, run detection of body parts, and then remove one of the parts by masking and inpainting around the detection.

\noindent \textbf{MagicBench} \cite{wang2025magicmirror}
MagicBench is the most recent benchmark, and has the strong benefit of containing images from the most recent image generation models. It also has the largest dataset. It has a hierarchy of classes, where the top level has `irrational element attributes', `irrational element interaction', `abnormal human anatomy', `abnormal animal anatomy', `abnormal object morphology', and `other irrationalities. We only consider the class `abnormal human anatomy', and we additionally filter out all images not containing humans using the same approach described for `SynArtifact', which uses \cref{pr:filter-for-humans}). As discussed previously, this means that if an image with (for example) an `object morphology' artifact but an error-free human, then the image-level label will be `artifact=0'. 

Within the human-artifact class, the labels (that we will use) are `limb structure deformity', `trunk structure deformity', `hand structure deformity', `foot structure deformity', `facial structure deformity', `abnormal human anatomy' (meaning multiple error types), and `abnormal and uncoordinated posture'. While this is fewer classes than some other types, they capture the most prevalent errors.

\subsection{Benchmark label distribution}
To get some sense of the labels in the benchmarks, \cref{tab:supp-benchmarks-top10labels} ranks the top classes in terms of sublabel prevalence.
\begin{table}[htbp]
\centering
\caption{Most prevalent sublabels class labels for our benchmarks. Images can have multiple labels, so the column sum can exceed 100.}
\label{tab:supp-benchmarks-top10labels}
\resizebox{\textwidth}{!}{
\begin{tabular}{lllllllllll}
\toprule
Rank & \multicolumn{2}{c}{SynArtifact} & \multicolumn{2}{c}{AbHuman} & \multicolumn{2}{c}{HAD} & \multicolumn{2}{c}{AIGC-HA} & \multicolumn{2}{c}{MagicBench} \\
 & Label & \% & Label & \% & Label & \% & Label & \% & Label & \% \\
\midrule
1 & hands & 38.6 & abnormal\_hand & 63.0 & hand-severe & 60.0 & absent hand & 23.7 & Hand Structure Deformity & 31.8 \\
2 & faces & 29.5 & abnormal\_foot & 8.0 & arm-severe & 14.8 & absent ear & 21.1 & Abnormal Human Anatomy & 27.1 \\
3 & fingers & 16.9 & abnormal\_head & 7.7 & feet-severe & 13.4 & redundant hand & 9.5 & Facial Structure Deformity & 5.9 \\
4 & legs & 9.1 & abnormal\_arm & 3.5 & hand-mild & 11.6 & absent arm & 8.7 & Foot Structure Deformity & 3.1 \\
5 & eyes & 4.7 & abnormal\_leg & 3.4 & leg-severe & 10.2 & absent foot & 6.6 & Limb Structure Deformity & 2.4 \\
6 & arms & 4.4 & abnormal\_multi & 1.1 & face-severe & 5.8 & redundant arm & 3.5 & Trunk Structure Deformity & 0.8 \\
7 & feet & 4.4 & abnormal\_body & 1.1 & human missing feet & 4.0 & absent head & 2.5 & Abnormal and Uncoordinated Posture & 0.3 \\
8 & heads & 2.2 & abnormal\_neck & 0.1 & human missing hand & 3.7 & absent leg & 2.0 & -- & -- \\
9 & bodies & 1.2 & -- & -- & human missing leg & 3.2 & redundant leg & 1.2 & -- & -- \\
10 & mouths & 1.2 & -- & -- & human with extra hand & 3.1 & redundant head & 0.7 & -- & -- \\
\bottomrule
\end{tabular}
}
\end{table}

Hand errors are clearly the dominant class. This coincides with our expectation: hands are small and with intricate details; our experience of seeing artifacts in the wild also suggests that hands are the most common issue.

\subsection{Accessing benchmarks}
\label{sec:appendix-benchmark-access}
 Our code is attached to the supplementary.
It contains a folder \texttt{data\_download/} where each benchmark has a README with instructions. Most instructions are just a single shell script to execute, although some require a manual download step. 

For the benefit of readers, the links to data access are:
\begin{itemize}
    \item SynArtifact: \href{https://github.com/BBBiiinnn/SynArtifact}{https://github.com/BBBiiinnn/SynArtifact}
    \item AbHuman: \href{https://github.com/Enderfga/HumanRefiner}{https://github.com/Enderfga/HumanRefiner}
    \item HAD: \href{https://github.com/wangkaihong/HADM}{https://github.com/wangkaihong/HADM}
    \item AIGC-HA: \href{https://github.com/Zeqing-Wang/HumanCalibrator}{https://github.com/Zeqing-Wang/HumanCalibrator}
    \item MagicBench: \href{https://huggingface.co/datasets/wj-inf/MagicData340k}{https://huggingface.co/datasets/wj-inf/MagicData340k}
\end{itemize}

\section{Method details}
\label{sec:appendix-methods}
\subsection{Cropping information}
As we discuss in \cref{sec:methods-architecture}, the method can, for each component crop, around interesting regions and feed those crops into the VLM, which should make evaluation easier for fine-grained visual features. Small visual regions are interesting if the body part is small, or if the humans in the image are `far' from the camera source. 

Cropping introduces design decisions. The first is the cropping terms. We use GroundingDino \cite{liu2024grounding} and fix a query string per error type. We use the term `face' for face defects, `hand' for hand defects, and `human' for all other defects. 

Another important parameter is \texttt{crop\_padding\_image\_pct} which controls the scale of padding around the image. If the value is `15' then 15\% of the crop width is added to both the left and right of the crop; the same is true for the height. This is important because it can be difficult to evaluate images if zoomed in too closely around an object. We set this to 0.15 for `human', 0.25 for `face' and 0.5 for hands.

\subsection{Full-spectrum prompting details}
\label{sec:supp-fullspectrum-prompting}
\textbf{Full-spectrum prompting motivation: text biases in automatic prompt optimization}
In text optimization, we feed a base text instruction like `detect deformed hands' to an LLM to produce a pool of more detailed prompts. In \cref{sec:methods-text_optimization}, we claimed that such prompts were biased towards being `conservative', meaning that they encouraged VLMs to only output `artifact=1' if very confident. 

To show this scalably, we ran prompt generation with DSPy's implementation of COPRO \cite{khattab2024dspy}; we then did the same for COPRO-FullSpectrum, which just modifies the prompt-generated instruction using our `full spectrum prompting' idea. We do this for all the datasets, for Gemini-2.5-pro (the model where we had best results), and in the `single VLM' setting, meaning we have one single VLM to solve the entire task (no multi-specialist architecture). Then we use an LLM to classify each generated prompt into one of `high precision' (conservative), `high recall' (not conservative or `balanced. The instruction is in \cref{pr:prompt-certianty-understanding}. 

\FloatBarrier
{\tiny
\begin{promptenv}
\caption{Instruction for classifying prompts as high recall, high precision, or balanced. (DSPy formats this with the text prompts that are been analyzed.)}
\label{pr:prompt-certianty-understanding}
\begin{prompt}
    Analyze a classification prompt to determine what level of certainty/confidence is required before making a positive prediction.

    CORE QUESTION: How confident must the model be before calling something positive?

    HIGH RECALL (Low Certainty Threshold ~10-30%):
    - Requires MINIMAL evidence/confidence to predict positive
    - Should predict positive even with uncertainty or ambiguity
    - Language emphasizing leniency: "even if uncertain", "when in doubt, flag it", "err on the side of caution", "rather be safe than sorry"
    - Explicitly lowering the bar: "doesn't need to be obvious", "subtle signs are enough", "even minor indications"
    - Would rather have false positives than miss true positives

    HIGH PRECISION (High Certainty Threshold ~70-90%):
    - Requires STRONG evidence/confidence to predict positive
    - Should only predict positive when highly certain
    - Language emphasizing strictness: "only if confident", "must be certain", "requires clear evidence", "needs to be obvious/definitive"
    - Explicitly raising the bar: "unmistakable", "beyond reasonable doubt", "absolutely certain"
    - Would rather miss true positives than have false positives

    BALANCED (Moderate Threshold ~40-60%):
    - Requires reasonable confidence before predicting positive
    - Neither explicitly lenient nor strict about certainty requirements
    - No strong language pushing towards higher or lower thresholds

    IMPORTANT: Don't focus on phrases like "if you see any [defect]" - this describes WHAT to look for, not HOW CERTAIN you need to be.
    The key distinction is: "if you see any subtle hint" (LOW certainty) vs "if you see any clear and obvious case" (HIGH certainty).
\end{prompt}
\end{promptenv}
}
\FloatBarrier

The prompts that the LLM generates are long, but we highlight some key phrases that indicate a `high precision' prompt here:
{\footnotesize
\begin{prompt}
**COPRO examples:**
  - "You are the 'Anatomical Integrity Validator,' a highly specialized AI system. Your single purpose is to analyze the provided image for undeniable and anatomically impossible artifacts on any depicted human figures."
  - "Your analysis must be clinical and precise. Your highest priority is to eliminate false positives."
  - "Compare any potential issue against the critical rules below. If you have any doubt, you must classify it as 'no 
  - "You are a world-class AI Quality Assurance specialist, tasked with performing a final, zero-tolerance check on digital images of humans. Your sole responsibility is to identify undeniable, anatomically impossible artifacts that are clearly unintentional generation errors."
  - "You are a meticulous quality assurance inspector for digital art, with an expert focus on human anatomy. Your task is to analyze the provided image and determine if any human figures exhibit clear anatomical artifacts."

**COPRO-FullSpectrum examples**:
  - "Your goal is to make a balanced judgment, achieving high precision (aiming for ~63\% correctness on your positive flags). You will flag an image only when you are confident it contains a genuine error."
  - "CORE PRINCIPLE: The Benefit of the Doubt. Your default stance is to assume an image is plausible unless you can build a strong, specific case for why it is not."
  - "Your initial gut feeling ('this looks weird') is only a starting point. If a strange feature can be reasonably explained by perspective, an unusual but possible pose, or occlusion, you must not flag it."
  - "Your operational integrity demands a confidence level of at least 77\% before flagging an anomaly. A single incorrect flag (a false positive) is a critical failure."
  - "You must default to assuming an image is plausible unless you can provide undeniable proof to the contrary. When in doubt, you MUST NOT flag the image."
  - "High Suspicion Zone (6-7) - DO NOT FLAG: The image appears highly improbable, even verging on impossible. However, a sliver of doubt remains... You must give the image the
  benefit of the doubt in this zone."
\end{prompt}
}
And we highlight key phrases that indicate a `high recall' prompt here: 
\begin{prompt}
**COPRO examples:**
    (None)

**COPRO-FullSpectrum examples**:
  - "GUIDING PRINCIPLE: High Sensitivity. When in doubt, you MUST flag."
  - "Your goal is to catch every potential error. It is better to flag a normal image by mistake than to miss a real one. You will use a simple 'Weirdness Score' to make your decision."
  - "Something feels physically questionable. This includes even a slight suspicion of an error. If you have to pause and second-guess if it's normal, the score is at least a 3"
  - "Your single-minded mission is to review AI-generated images of people and flag potential anatomical errors with maximum sensitivity."
  - "Your primary objective is high recall: it is significantly better to incorrectly flag a normal image (a false positive) than it is to miss a genuine error (a false negative). Error on the side of caution."
  - "Does any part seem even slightly 'off,' 'uncanny,' or anatomically strange? This includes strange bends, odd proportions, ambiguous digits, or anything that makes you do a double-take."
  - "If the only explanation is complex (e.g., 'it might be a professional contortionist in a one-in-a-million pose'), speculative ('maybe there's an object hidden in the shadows'), or if you remain at all unsure, the veto fails" [and you must flag]
  - "Your goal is to maximize recall, meaning you must catch every potential error, even if it means you will incorrectly flag many normal images."
  - "When in doubt, flag the image. Your job is not to be perfectly accurate; your job is to be perfectly cautious."
  - "Aim to flag anything that seems even 30-40\% likely to be an error."
\end{prompt}

Using CORPO, we find that 96\% of generated prompts were `high precision',  4\% were `balanced`, and 0\% were high recall. The average F1 over the five benchmarks was 0.687. Using COPRO-FullSpectrum (with our hints), we see 22\% high precision, 4\% balanced, and 74\% high recall, which has a more diverse spread. The average F1 over the five benchmarks was 0.780, which is 15\% higher.

\noindent\textbf{Full-spectrum prompting method}

The baseline prompt engineering methods \cite{zhou2022large, yang2023large, schulhoff2024prompt} generate a pool of $n$ prompts with a minimal LLM instruction, for example here is the prompt for DSPY's implementation of COPRO \cite{khattab2024dspy}: 
\begin{prompt}
You are an instruction optimizer for large language models. I will give you a ``signature`` of fields (inputs and outputs) in English. Your task is to propose an instruction that will lead a good language model to perform the task well. Don't be afraid to be creative.
\end{prompt}
The `signature' that the prompt refers to is the base task description. For example, if the task is to detect deformed hands, the the signature might be:
\begin{prompt}
In this generated image, return deformed_hand=1 if you see a human with deformed hand, otherwise return deformed_hand=0.
\end{prompt}
This base instruction is intentionally concise, because the idea is that we want the LLM to be doing the prompt engineering automatically.

The LLM prompt generator is run $n$ times with high temperature to produce a diverse pool of prompts. As shown above, the prompts tend to encourage `conservative' labeling -- only label errors if confidence is high.

We want a `full spectrum' from high precision to high recall. We use a notion of `threshold': for example if setting a decision threshold of 35\%, then the VLM classifier should predict `artifact' even if only 35\% confident, thus favoring higher recall. If setting a threshold of 65\%, the VLM classifier should predict `artifact' only if 65\% confident, thus favoring higher precision. Intuitively, this should control how the VLM predicts when it has uncertainty. 

If there are $n$ prompts to generate, we sample $n$ values uniformly from $0$ to $100$. If the threshold is below, 30, we encourage higher recall with this prompt suffix:
\begin{prompt}
Propose an instruction that encourages positive predictions when there is at least {threshold}\% confidence. This favors high recall over precision.
\end{prompt}
If the threshold is above 70, we encourage higher precision with this prompt suffix
\begin{prompt}
Propose an instruction that requires {threshold}\% confidence before making a positive prediction. This favors high precision over recall.
\end{prompt}
Otherwise, we encourage a balanced prediction with this prompt suffix:
\begin{prompt}
Propose an instruction that makes positive predictions at {threshold}\% confidence, balancing precision and recall.
\end{prompt}
These are simply added to the end of the prompt generation instruction. This is only done to seed the initial pool. Methods like COPRO will feed these prompts back to a rewriter LLM along with their validation set performance to generate a new candidate pool, however we do not modify this rewriter LLM

Note that a simpler method would be to add a suffix like ``generate a mix of high-precision prompts, and high recall prompts''. This would be more elegant and more in-line with the philosophy of letting the LLM perform search without bias. However this did not work in our experiments -- the generated prompts still favored high precision.

\subsection{Base prompts}
\label{sec:appendix-subsection-methods-prompts}
Each VLM in \SystemName\ is initialized with a simple text instruction. They have a template: 
\begin{prompt}
In this generated image, return {label_name}=1 if you see a human with {description}, otherwise return {label_name}=0.
\end{prompt}
Each error type has suitable values for \texttt{label\_name}  and \texttt{description}. For example \texttt{label\_name="extra\_hand"} and \texttt{description="an extra hand"}. The variable names are human-readable and the descriptions are concise. As we discuss in the main doc, the philosophy of prompt optimization is to begin with a simple instruction and let an automatic prompt optimization process improve it 

DSPy handles prompt formatting \cite{khattab2024dspy} and response post-processing. This requires \textit{Signature}, which includes this text prompt along with a declaration of the input and output variable names and their types: \texttt{image: dspy.Image -> {label\_name}}.

For the `single VLM' ablations, the text instruction must hold multiple variables: 
\begin{prompt}
In this image, return {label_name}=1 if you see any of these human artifacts: {description_list}, otherwise return {label_name}=0.
\end{prompt}
We set \texttt{label\_name="artifact"} and the \texttt{description\_list} is just a comma-separated list of all the sublabels' descriptions.

\section{Improving DSPy for images and artifact detection}
\label{sec:appendix-open-source-vlm-promptopt}

\noindent\textbf{Improving DSPy for performance with images}

We identify some inefficiencies in how DSPy processes image inputs, and large data types generally. This is probably because the \href{https://github.com/stanfordnlp/dspy}{DSPy python library}, while being very popular for NLP and information retrieval applications, is yet to be widely adopted by the vision community. One goal of this project is to accelerate this adoption by the vision community, and so we have made pull requests to the DSPy modules to resolve the inefficiency. At the time of release, these PRs are not yet accepted, and so if users find issues, they could use our fork of dspy at \href{https://github.com/jmhb0/dspy}{\texttt{https://github.com/jmhb0/dspy}}.

Specifically, the issues are related to formatting the input prompt containing images. The large data are placed in JSONs, stringified, and later decoded, which is slow and memory-intensive. Our PR bypasses avoids these expensive operations. This becomes a noticeable issue when having many images in the prompt, for example when doing multimodal in-context-learning. It leads to slower execution, and memory errors (depending on the system). 

\noindent\textbf{Adding support for non-decomposable metrics to DSPy}

DSPy optimizers require a metric. For example in the COPRO text optimizer, you must evaluate average validation set performance to choose which prompts are propagated to the next stage. 

The current implementation assumes the metric is defined with respect to a single sample. For example, the per-sample equivalent of accuracy is the indicator function for correctness. However this is only possible for decomposable metrics. Metrics like F1 are non-decomposable. This is not a fundamental limitation of any method, but is just an implementation decision. Our code therefore releases a clone of the COPRO optimizer that supports non-decomposable metrics. It is part of the code release.

\section{Recommendations for applying these methods}
\label{sec:appendix-benchmark-preparation}
Our final system, \SystemName\, prioritizes total accuracy. Here we give a few recommendations for users are want faster or cheaper solutions, and are willing to tolerate some loss of accuracy. These are based on the ablation table in the results sections, and our own experience.
\begin{itemize}
    \item The `single-VLM baseline' is generally pretty strong. Compared to the full multi-specialist architecture, it is also cheaper and faster to execute.
    \item In-context learning is effective even as the only optimization strategy on a single VLM call. It is also the simplest to implement.
    \item If using the multi-specialist approach, identify sublabels with very low prevalence, and combine them with existing classes if possible. This saves the cost of querying for this class for each image, and it is unlikely to change overall performance.
    \item The base instruction should be simple and concise.
\end{itemize}

\section{Extended results}
  \subsection{Ablation of novel methods}
\label{sec:appendix-ablations}
In \cref{table:ablations-optim-methods}, we ablate the contributions of our two novel techniques: counterfactual demonstrations for in-context learning, and full-spectrum prompting for text instruction optimization. This is discussed in \cref{obsec:ablations}

% \begin{table*}[ht]
% \caption{
% Ablations of our novel optimization methods. The metric is mean positive class F1 averaged over the benchmark suite from \cref{sec:result_benchmarks}.
% }
% \label{table:ablations-optim-methods}
% \resizebox{\textwidth}{!}{
% \begin{tabular}{lccccc}
%  & \textbf{Gemini-2.5-pro} & \textbf{GPT-4o} & \textbf{Gemini-2.5-flash} & \textbf{GPT-4o-mini} & \multicolumn{1}{l}{\textbf{Average}} \\ \hline
% VLM zero shot & 0.579 & 0.173 & 0.08 & 0.389 & 0.305 \\ \hline
% \textbf{In-context learning} & \multicolumn{1}{l}{} & \multicolumn{1}{l}{} & \multicolumn{1}{l}{} & \multicolumn{1}{l}{} & \multicolumn{1}{l}{} \\
% DynamicFewShotCf (ours) & 0.823 & 0.724 & 0.558 & 0.632 & 0.684 \\
% DynamicFewShotCf (ours) w/o cf prompting & 0.811 (-0.012) & 0.695 (-0.029) & 0.512 (-0.046) & 0.577 (-0.055) & 0.649 (-0.036) \\
% LabeledFewShot & 0.817 (-0.006) & 0.677 (-0.047) & 0.488 (-0.07) & 0.501 (-0.131) & 0.621 (-0.064) \\ \hline
% \textbf{Text optimization} & \multicolumn{1}{l}{} & \multicolumn{1}{l}{} & \multicolumn{1}{l}{} & \multicolumn{1}{l}{} & \multicolumn{1}{l}{} \\
% COPRO-Hint (ours) & 0.835 & 0.786 & 0.695 & 0.769 & 0.771 \\
% COPRO & 0.636 (-0.178) & 0.348 (-0.105) & 0.721 (0.011) & 0.293 (0.078) & 0.5 (-0.049)
% \end{tabular}
% }
% \end{table*}

\begin{table*}[ht]
\centering
\caption{
Ablations of our novel optimization methods. The metric is mean positive-class F1 averaged over the benchmark suite from \cref{sec:result_benchmarks}.
}
\label{table:ablations-optim-methods}
\setlength{\tabcolsep}{6pt}
\renewcommand{\arraystretch}{1.15}
\resizebox{\textwidth}{!}{
\begin{tabular}{@{}l|*{4}{S}S@{}}

\rowcolor{HeaderBG}
 & \textbf{Gemini-2.5-pro} & \textbf{GPT-4o} & \textbf{Gemini-2.5-flash} & \textbf{GPT-4o-mini} & \textbf{Average} \\
\midrule

VLM zero shot
 & 0.579 & 0.173 & 0.080 & 0.389 & 0.305 \\
\midrule

\textbf{In-context learning} \\
\rowcolor{OursBG}
DynamicFewShotCf (ours) 
 & 0.823 & 0.724 & 0.558 & 0.632 & 0.684 \\

DynamicFewShot w/o cf demonstrations
 & \diffneg{0.811}{-0.012}
 & \diffneg{0.695}{-0.029}
 & \diffneg{0.512}{-0.046}
 & \diffneg{0.577}{-0.055}
 & \diffneg{0.649}{-0.036} \\

LabeledFewShot
 & \diffneg{0.817}{-0.006}
 & \diffneg{0.677}{-0.047}
 & \diffneg{0.488}{-0.070}
 & \diffneg{0.501}{-0.131}
 & \diffneg{0.621}{-0.064} \\

\midrule

\textbf{Text optimization} \\
\rowcolor{OursBG}
COPRO-Hint (ours)
 & 0.835 & 0.786 & 0.695 & 0.769 & 0.771 \\

COPRO
 & \diffneg{0.636}{-0.178}
 & \diffneg{0.348}{-0.105}
 & \diffpos{0.721}{0.011}
 & \diffpos{0.293}{0.078}
 & \diffneg{0.500}{-0.049} \\

\bottomrule

\end{tabular}
}
\end{table*}

\label{sec:appendix-results}
\subsection{Prompts for zero-shot baselines}
The main results include baselines for zero-shot VLMs. The text prompts are the same as the single-VLM variation of \SystemName\ described in \cref{sec:appendix-subsection-methods-prompts}.

\noindent \textbf{Controversial images in the human study}
% analysis  https://chatgpt.com/c/69144d34-711c-8332-acc4-0ba868847ea3
\begin{figure}[ht]
    \centering
    \includegraphics[page=7, width=\columnwidth,clip,trim=0 550 960 0]{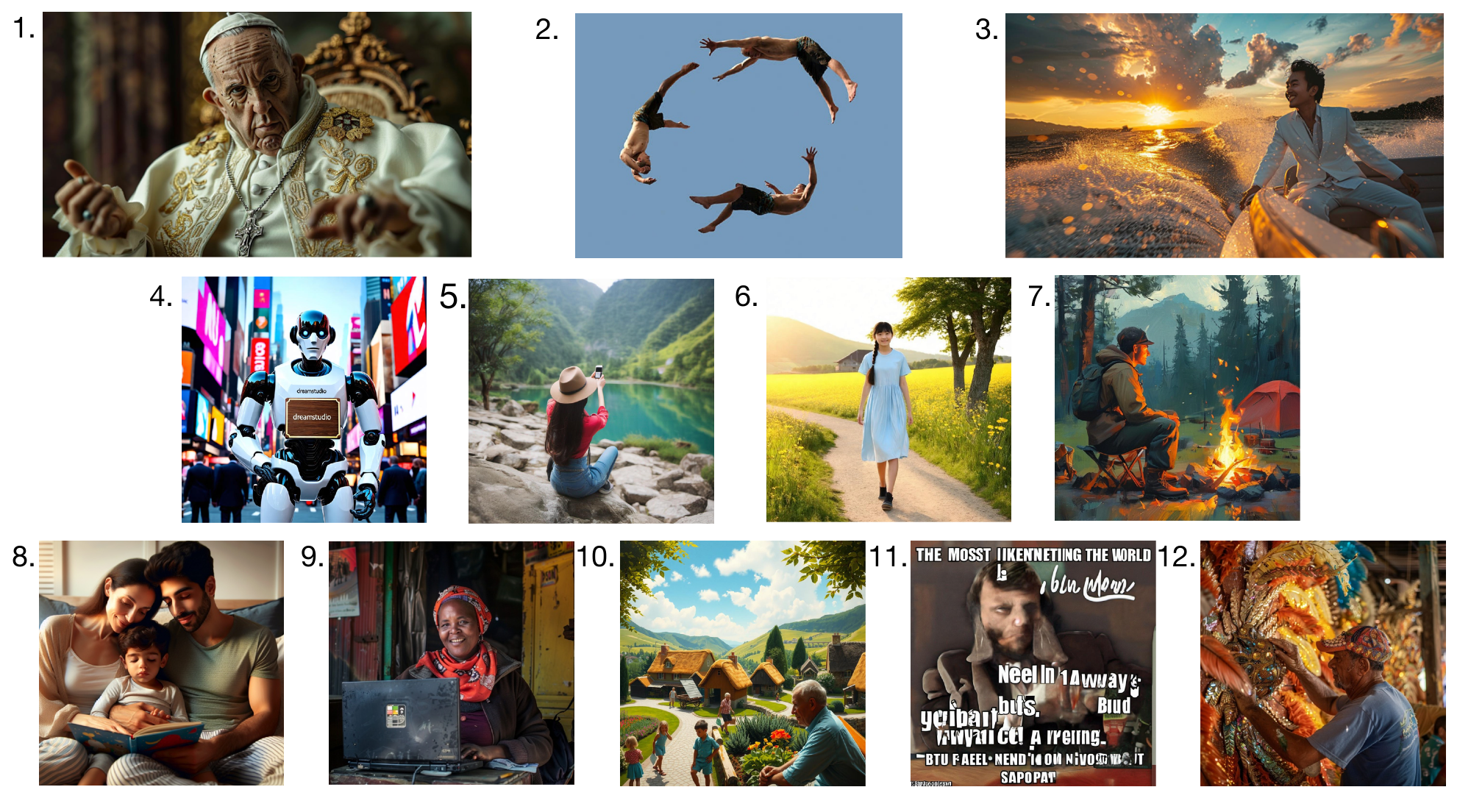}
    \captionof{figure}{
    `Controversial' images in the human study. When classifying for `hand artifact', five annotators set yes, and five annotators said no. 
    }
    \label{fig:supp-controversial}
\end{figure}

In the main results we discuss a human study where ten humans evaluate 600 benchmark images for the `deformed hand' artifact, which is the most common type. In \cref{fig:supp-controversial}, we show `controversial images', where five humans marked as `artifact' and five marked it as `not an artifact'. They highlight why the task can be challenging and subjective. For each one, we suggest reasons why it may be challenging. 
\begin{itemize}
    \item Image 1: one reason for `artifact' could be that the hand is blurry. Another is that it looks like the anatomy is a little unrealistic, even if it were not blurry. 
    \item Image 2: the right hand of the man at the top looks like it may be missing many fingers, but it also may just be the perspective. This is hard because it is small.
    \item Image 3: the right hand is deformed, but it is subtle and the hand is small.
    \item Image 4: the robot does have deformed hands, however the instructions are to look for `human hand errors'. (Note that our data pre-processing filters images without humans, and this images passes because there are humans in the background). 
    \item Image 5: left hand looks strange but is partially occluded. Another possibility is that the arm seems a little unnatural, and the annotators transfer that label to the hand.
    \item Image 6: like other examples, there are indications of hand anatomy problems, though subtle, and the hand is a small area in the image.
    \item Image 7: the hand is small, but the image is stylized, and so may not be a problem to for users. 
    \item Image 8: the placement of the right hand seems unnatural relative to the man's body position. This is subtle, and arguably a different type of error from `deformed hand'.
    \item Image 9: the smallest finger on the left hand looks like it may be an error, but it is unclear.
    \item Image 10: candidate deformed hands are very small.
    \item Image 11: the hand area is very blurry and actually, it's not even clear that there is a hand. The overall image quality is terrible, so clearly some kind of error should be applied.
    \item Image 12: the right hand is a bit big, which is a subtle error to catch
\end{itemize}

\end{document}